\documentclass[lettersize,journal]{IEEEtran}
\usepackage{amssymb}
\usepackage{algorithm}
\usepackage{bibentry}
\usepackage{subfigure}
\usepackage{multirow}
\usepackage{epsfig}
\usepackage{color}
\usepackage{epstopdf}
\usepackage{rotating}
\usepackage{float}
\usepackage{multicol}

\usepackage{bbding}
\usepackage{balance}
\usepackage{microtype}
\usepackage{makecell}
\usepackage{mathtools}
\usepackage{etoolbox}
\usepackage{cases}
\usepackage{enumitem}
\usepackage{xcolor}

\usepackage{tabularx}
\usepackage{threeparttable}
\usepackage{soul}

\usepackage{amsmath,amsfonts}
\usepackage{algorithmic}
\usepackage{array}
\usepackage[caption=false,font=normalsize,labelfont=sf,textfont=sf]{subfig}
\usepackage{textcomp}
\usepackage{stfloats}
\usepackage{url}
\usepackage{verbatim}
\usepackage{graphicx}
\usepackage{booktabs}
\usepackage{color}

\def\BibTeX{{\rm B\kern-.05em{\sc i\kern-.025em b}\kern-.08em
    T\kern-.1667em\lower.7ex\hbox{E}\kern-.125emX}}

\begin{document}
\title{Large Language Models are In-Context Molecule Learners}
\author{Jiatong Li, Wei Liu, Zhihao Ding, Wenqi Fan, Yuqiang Li, and Qing Li

\IEEEcompsocitemizethanks{
\IEEEcompsocthanksitem J. Li, Z. Ding, W. Fan, and Q. Li are with the Department of Computing, The Hong Kong Polytechnic University. E-mail:  \{jiatong.li, tommy-zh.ding\}@connect.polyu.hk, wenqifan03@gmail.com, csqli@comp.polyu.edu.hk. 
\IEEEcompsocthanksitem W. Liu is with Shanghai Jiao Tong University. E-mail: captain.130@sjtu.edu.cn.
\IEEEcompsocthanksitem Y. Li is with Shanghai AI Lab. E-mail: liyuqiang@pjlab.org.cn.

}
\thanks{(Corresponding authors: Wenqi Fan, Yuqiang Li, and Qing Li.)}
}

\markboth{IEEE TRANSACTIONS ON KNOWLEDGE AND DATA ENGINEERING, SUBMISSION 2024}%
{Shell \MakeLowercase{\textit{et al.}}: Bare Demo of IEEEtran.cls for Computer Society Journals}

\maketitle

\begin{abstract}
Large Language Models (LLMs) have demonstrated exceptional performance in biochemical tasks, especially the molecule caption translation task, which aims to bridge the gap between molecules and natural language texts.
However, previous methods in adapting LLMs to the molecule-caption translation task required extra domain-specific pre-training stages, suffered weak alignment between molecular and textual spaces, or imposed stringent demands on the scale of LLMs.
To resolve the challenges, we propose \textbf{I}n-\textbf{C}ontext \textbf{M}olecule \textbf{A}daptation (\textbf{ICMA}), as a new paradigm allowing LLMs to learn the molecule-text alignment from context examples via In-Context Molecule Tuning. 
Specifically, ICMA incorporates the following three stages: Hybrid Context Retrieval, Post-retrieval Re-ranking, and In-context Molecule Tuning.
Initially, Hybrid Context Retrieval utilizes BM25 Caption Retrieval and Molecule Graph Retrieval to retrieve similar informative context examples. 
Additionally, Post-retrieval Re-ranking is composed of Sequence Reversal and Random Walk selection to further improve the quality of retrieval results. 
Finally, In-Context Molecule Tuning unlocks the in-context learning and reasoning capability of LLMs with the retrieved examples and adapts the parameters of LLMs for better alignment between molecules and texts.
Experimental results demonstrate that ICMA can empower LLMs to achieve state-of-the-art or comparable performance without extra training corpora and intricate structures, showing that LLMs are inherently in-context molecule learners.
\end{abstract}


\section{Introduction}
Molecules play a crucial role across various fields, such as medicine~\cite{ding2019selective}, agriculture~\cite{twyman2003molecular}, and material science~\cite{higuchi2023material}, as they are widely used in the development of drugs, fertilizers, and advanced materials. Recently, LLMs have demonstrated remarkable success in the molecular domain, as molecules can be represented as Simplified Molecular-Input Line-Entry System (SMILES) strings~\cite{weininger1988smiles}, which can be comprehended and generated by LLMs in a similar manner to natural languages. 
To further bridge the gap between the molecules and natural languages, MolT5~\cite{edwards-etal-2022-translation} proposes the molecule-caption translation task, which comprises two sub-tasks: molecule captioning (Mol2Cap) and text-based de novo molecule generation (Cap2Mol). Specifically, Mol2Cap involves generating a textual description that elucidates the features of the given molecule, while Cap2Mol focuses on predicting the exact molecule based on the textual caption. The study of the molecule-caption translation task offers an accessible and chemist-friendly venue for molecule discovery, which has raised wide research focus.
\begin{figure}
    \centering
    \includegraphics[width=\columnwidth]{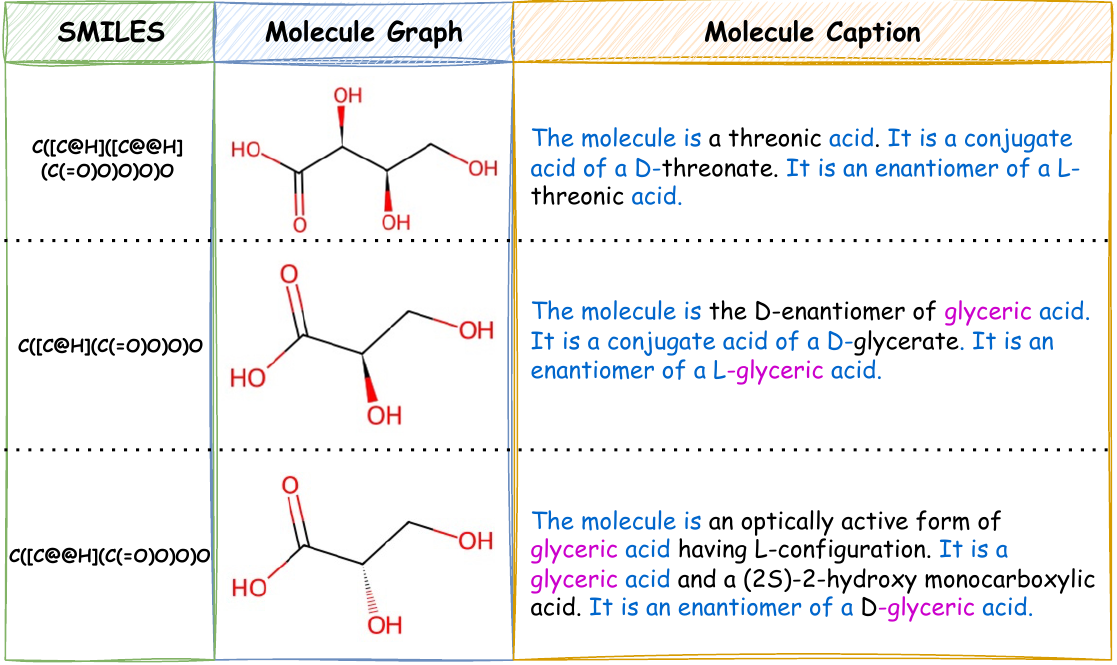}
    \caption{An illustration of three similar molecules alongside their molecule captions. The molecules are represented as both SMILES strings and graphs, while the molecule captions elucidate their structures and functions. Here, the three molecules are similar, considering their 2D graph embeddings, and the overlaps in their captions are highlighted in blue and pink.}
    \label{fig:intro}
\end{figure}

Generally, there are two main paradigms for adapting LLMs to the molecule-caption retrieval task. 
The first paradigm is the domain-specific pre-training \& fine-tuning. For instance, MolT5~\cite{edwards-etal-2022-translation} first proposes and handles the molecule-caption translation task as the language translation task, pre-training the MolT5 model with chemical corpora like PubChem~\cite{kim2019pubchem} and then fine-tuning the model on the ChEBI-20 dataset~\cite{edwards2021text2mol}. Additionally, MoMu~\cite{su2022molecular} and MolCA~\cite{liu2023molca} introduce an extra modality alignment stage before fine-tuning on downstream tasks, which aligns the output of 2D molecule graph encoder with the input space of LLMs.
In contrast, the other paradigm involves prompting and utilizing the in-context learning capability of LLMs. For example, MolReGPT~\cite{li2023empowering} introduces In-Context Few-Shot Molecule Learning, prompting general LLMs like GPT-3.5 and GPT-4 to achieve competitive performance without extra parameter adaptations.

However, the current paradigms face critical challenges. \textbf{On one hand}, the domain-specific pre-training \& fine-tuning paradigm requires extra pre-training stages (i.e., domain-specific pre-training and modality alignment), which is challenging due to the scarcity of high-quality chemical datasets, especially molecule-caption pairs, making this paradigm infeasible to scale up to the most advanced LLMs with billion parameters.
Besides, the domain-specific pre-training \& fine-tuning paradigm also suffers from weak alignment between molecules and texts, as phrases in molecule captions often indicate specific sub-structures of molecules rather than the entire molecule.
Despite attempts to introduce extra modalities for better alignment \cite{su2022molecular, liu2023molca}, the integration of the additional modalities (e.g., 2D molecule graph) is still focused on the entire graph level and can only be applied to the Mol2Cap task, while ignoring the text-based generation of molecules, which is much more valuable for drug discovery.
\textbf{On the other hand}, the in-context learning \& prompting paradigm puts a harsh requirement on LLMs' emergent capabilities, such as reasoning and in-context learning abilities. However, LLMs with these emergent capabilities usually have billions of parameters, making them computationally expensive.
Consequently, there is a demand for a unified and efficient approach that effectively enhances the performance of the most advanced LLMs in both two sub-tasks of molecule-caption translation.

In this case, we propose \textbf{I}n-\textbf{C}ontext \textbf{M}olecule \textbf{A}daptation (\textbf{ICMA}) as a new paradigm for adapting LLMs in molecule-caption translation. 
Different from previous paradigms, ICMA aims to instruct LLMs to derive knowledge from informative context examples, especially the alignment between molecule SMILES representations and captions, via In-Context Molecule Tuning. As shown in Figure \ref{fig:intro}, similar molecules often share similar properties, as indicated by the overlaps among molecule captions. Conversely, similar captions tend to describe molecules with similar SMILES representations. In this case, with ICMA, general LLMs could fulfill their reasoning and in-context learning capability to better grasp the alignment between molecules and textual captions from context examples, thereby achieving better performance.

Specifically, ICMA incorporates three stages: Hybrid Context Retrieval, Post-retrieval Re-ranking, and In-context Molecule Tuning.
In the initial stage, Hybrid Context Retrieval, we employ Caption Retrieval and Molecule Graph Retrieval to fetch similar captions and molecules, respectively.
Subsequently, we introduce the Post-retrieval Re-ranking stage to enhance the quality of the retrieval algorithms. This stage incorporates two innovative strategies: Sequence Reversal and Random Walk, which aim to refine and reprioritize the retrieved examples.
Finally, we apply In-context Molecule Tuning to adapt the parameters of LLMs, enabling them to learn from the contextual mappings and effectively ground the current generation task.
Experiments are conducted across two real-world molecule-caption translation datasets, ChEBI-20 and PubChem324k. Results show that ICMA could enable LLMs to achieve state-of-the-art or comparable performance in both the two sub-tasks (i.e., Mol2Cap and Cap2Mol). Meanwhile, we also study the factors related to the model performance, including retrieval algorithms, context settings (i.e., context example number and maximum input length), model scales, and backbone LLMs. Lastly, the ablation study and detailed case study are conducted to justify the effectiveness of Post-retrieval Re-ranking components.

Our contribution mainly lies in:
\begin{itemize}
    \item We propose In-context Molecule Adaptation (ICMA) to improve the performance of LLMs in the molecule-caption translation task. ICMA could empower the reasoning and in-context learning capabilities of LLMs for better alignment between molecules and texts.
    \item We implement ICMA through three stages, including Hybrid Context Retrieval, Post-retrieval Re-ranking, and In-Context Molecule Tuning, significantly enhancing the informativeness of context examples.
    \item We conduct synthetic experiments, and the results show that our method enables LLMs to outperform previous paradigms, enabling better alignment between molecules and texts. Notably, our approach elevates general LLM like Mistral-7B to establish superior performance across both the two sub-tasks of molecule-caption translation, achieving 0.581 BLEU-4 score in Mol2Cap and 0.460 exact-matched score in Cap2Mol. Additionally, we comprehensively study the mechanism and influential factors of ICMA, showing that LLMs are inherently in-context molecule learners.
\end{itemize}

\section{Related Work}
\label{sec:relatedwork}
In this section, we discuss the related work of molecule-caption translation and the development of in-context learning.

\subsection{Molecule-Caption Translation}
Inspired by the image captioning task, Edwards et al.~\cite{edwards2021text2mol} introduce a new dataset, ChEBI-20, with pairs of molecules and manually labeled captions that describe the molecular properties. The molecule-caption translation task was initially proposed in MolT5~\cite{edwards-etal-2022-translation}. Meanwhile, MolT5 proposes a T5 model that is jointly pre-trained on molecule SMILES and general text corpus.
MolXPT~\cite{liu2023molxpt} pre-trains a GPT model by introducing extra-wrapped texts as the pre-training corpus, demonstrating better molecule-text alignment.
However, the generation of SMILES strings suffers from the problem of invalid SMILES due to the mismatches of brackets.
To overcome the generation issue of SMILES strings, BioT5~\cite{pei2023biot5} introduces Self-referencing Embedded Strings (SELFIES)~\cite{krenn2020self} instead of SMILES strings to represent molecules in LLMs and proposes the BioT5 model that is jointly trained on single-modal data, wrapped text data, and molecule/protein-description pairs.
Meanwhile, as molecules can also be represented as graphs, some methods focus on molecule understanding by introducing molecule graph information.
For example, MoMu~\cite{su2022molecular} first proposes a graph encoder to encode molecule graph information and utilizes contrastive learning to bridge the semantic gap between the graph encoder and the LLM.
To better fuse the molecule graph information, MolCA~\cite{liu2023molca} follows the BLIP-2~\cite{li2023blip} and utilizes a Q-Former to project the output of the graph encoder into the LLMs, showing better molecule understanding performance.
However, most of these methods still adhere to the pre-training \& fine-tuning paradigm, which necessitates the joint pre-training of LLMs on both general text and extra chemical domain corpora. As the size of the training corpora and model weights of LLMs continue to increase, this approach has become extremely inefficient.
To address this issue, MolReGPT~\cite{li2023empowering} proposes the In-Context Few-Shot Molecule Learning to enable LLMs to learn the molecule-caption translation task from the context examples without modifying the model weights while still achieving comparable performance to these fine-tuned methods.

\subsection{In-Context Learning}
With the scaling of model size and corpus size~\cite{brown2020language, chowdhery2023palm}, LLMs emerge the in-context learning capability~\cite{wang2022self}, which enables LLMs to learn from contexts augmented with several examples~\cite{dong2022survey}. 
By utilizing the capability of ICL, LLMs can solve complex tasks without the necessity of being fine-tuned. 
For instance, with a few examples, GPT-3 could demonstrate similar performance to fine-tuned models in unseen tasks~\cite{brown2020language}.
What's more, based on context examples, LLMs could achieve better mathematical reasoning ability with the assistance of chain-of-thought (CoT)~\cite{wei2022chain}.
With the powerful in-context learning capabilities, Edwards et al.~\cite{edwards2023synergpt} propose in-context drug synergy learning to apply LLMs for personalized drug synergy prediction and drug design, while Jablonka et al.~\cite{jablonka2024leveraging} fine-tunes LLMs such as GPT-3 with chemical questions and answers to solve predictive tasks
In the scenario of molecule-caption translation, MolReGPT~\cite{li2023empowering} proves the retrieval quality is closely related to the model performance, while it still imposes strict requirements on the model scale, as the in-context learning capability only becomes apparent when the model weights reach a certain size.
\section{In-Context Molecule Adaptation}
\label{sec:methodlogy}
\begin{figure*}[htb]
    \centering
    \vskip -0.1in
    \includegraphics[width=\textwidth]{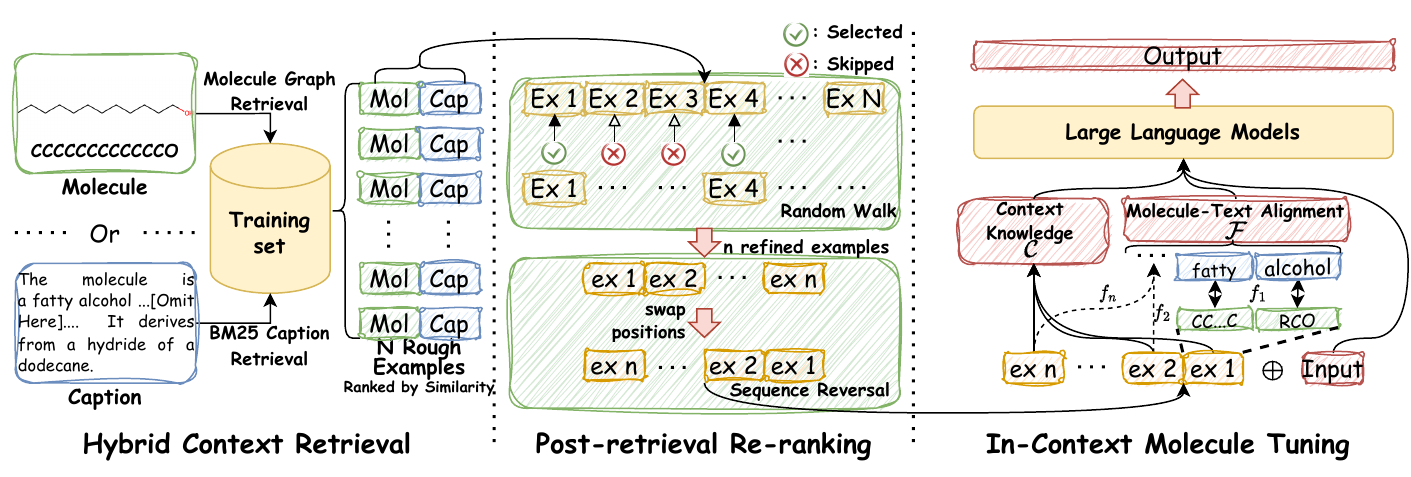}
    \caption{Framework of In-Context Molecule Adaptation (ICMA). Generally, ICMA consists of three stages, Hybrid Context Retrieval, Post-retrieval Re-ranking, and In-Context Molecule Tuning. 
    }
    \label{fig:model}
    \vskip -0.1in
\end{figure*}

In this section, we introduce In-Context Molecule Adaptation (ICMA) as a novel paradigm to adapt LLMs to molecule-caption translation. As shown in Figure~\ref{fig:model}, ICMA incorporates three stages, including Hybrid Context Retrieval, Post-retrieval Re-ranking, and In-context Molecule Tuning. Specifically, Hybrid Context Retrieval first retrieves $N$ rough examples from the training set $\mathcal{D}$ by calculating the similarity between the current query and the molecule-caption pairs from the training set. After that, Post-retrieval Re-ranking with Random Walk and Sequence Reversal is adopted to obtain $n$ refined examples from the $N$ rough examples. Finally, In-Context Molecule Tuning can be performed to update the parameters of LLMs to learn the molecule-text alignment from refined context examples. 

\subsection{Hybrid Context Retrieval}
The retrieval quality is closely related to the informativeness of context examples. For example, if the retrieved molecules are more similar to the current query molecule, they are likely to exhibit more overlaps in their respective caption, which could enable better alignments between molecules and texts. Therefore, the development of retrieval algorithms plays a crucial role in ICMA. In this work, we introduce Hybrid Context Retrieval, which adopts hybrid modalities (i.e., 2D molecule graph and text), as well as hybrid retrieval algorithms designed for the specific tasks (i.e., Molecule Graph Retrieval for Mol2Cap and BM25 Caption Retrieval for Cap2Mol). 

For the \textbf{Mol2Cap} task, ICMA adopts Molecule Graph Retrieval to better refine the retrieval quality.
Previously, MolReGPT~\cite{li2023empowering} utilizes the Morgan Fingerprints (Morgan FTS) and Dice similarity to evaluate the similarity between molecules, which encodes the pre-defined handcraft structures as the embedding of the molecule. However, Morgan FTS are typically based on handcrafted chemical feature extraction, which may have limited capability to capture the comprehensive information of complex molecule structures and properties. On the other side, Graph Neural Networks (GNNs) have been widely used to capture topological structures ~\cite{liu2022hs}. Pre-trained on millions of molecule graphs, GNNs could better understand the molecular structure, providing complete chemical semantics~\cite{ijcai2023p760}. This makes GNNs a better option for molecule similarity calculation.
In ICMA, we adopt a pre-trained GNN encoder to obtain the molecule graph embeddings:
\begin{align}
    \textbf{e}_m = G\!N\!N(g_m),
\end{align}
where $\textbf{e}_m$ denotes the embedding of the given 2D graph $g_m$ of the molecule $m$. Specifically, we adopt Mole-BERT~\cite{xia2022mole} as the GNN encoder.

Subsequently, cosine similarity is leveraged to evaluate the similarity between the current query molecule graph $m^q$ and the other molecule graphs $m^i$ in the training set of molecules ($m^i \in \mathcal{D}_m$). Thus, the molecule similarity ranking function $\mathcal{R}^m$ can be represented as:
\begin{align}
    \mathcal{R}^m(m^q, m^i) = \cos{(\textbf{e}_{m^q}, \textbf{e}_{m^i})}.
\end{align}
In the \textbf{Cap2Mol} task, we inherit the BM25 Caption Retrieval~\cite{robertson2009probabilistic} from MolReGPT~\cite{li2023empowering} as it focuses on the detail matching of molecule captions, showing competitive performance and is much faster than LLM-based methods like Sentencebert~\cite{reimers2019sentence}. Specifically, given captions in the test set as query captions $Q_c$ and the training set of captions $\mathcal{D}_c$, the caption similarity ranking function $\mathcal{R}^c$ can be denoted as:

{\small
\begin{align}
    \mathcal{R}^c\!(\!Q_c\!,\!\mathcal{D}_c\!)\!=\!\sum_{i=1}^T\!I\!D\!F(\!c^q_i\!)\!*\!\frac{t\!f(c^q_i,\mathcal{D}_c)*(k_1+1)}{t\!f(\!c^q_i,\!\mathcal{D}_c\!)\!+\!k_1\!*\!(\!1\!-\!b\!+\!b\!*\!\frac{|\mathcal{D}_c|}{avgdl}\!)},
\end{align}
}

\noindent where $T$ is the number of query terms in the query caption, $c^q_i$ is the $i$-th query term, $I\!D\!F(c^q_i)$ is the inverse document frequency of $c^q_i$, $t\!f(c^q_i, \mathcal{D}_c)$ is the term frequency of $c^q_i$ in $\mathcal{D}_c$, $k_1$ and $b$ are hyperparameters, $|\mathcal{D}_c|$ is the length of $\mathcal{D}_c$, and $avgdl$ is the average caption length in the corpus.

\subsection{Post-retrieval Re-ranking}
Although refined retrieval algorithms could bring better retrieval quality, there are still some problems considering the arrangement of context examples. Thus, we propose Post-retrieval Re-ranking with Random Walk and Sequence Reversal to re-rank the priorities and positions of context examples, thus enhancing the quality of In-Context Molecule Tuning.

\subsubsection{Random Walk}
The molecules ranked top by retrieval algorithms can sometimes share too many overlaps, impairing the informativeness of context examples. 
In this case, for the context diversity and generalization performance of ICMA, it is necessary to give these less similar examples a chance to be visited.
Inspired by the Random Walk mechanism in graph theory, we propose Random Walk as a post-retrieval method to select examples from the top-$N$ retrieved results so that examples with lower rank still have a chance to be selected, which provides more useful information and complements the context diversity.
Mathematically, we adopt a dynamic chance for examples with different ranks that gradually decays as the rank moves down. Specifically, for the $j$-th example in the $N$ rough results, where $1\leq j \leq $N, the possibility of skipping $p(j)$ is represented as:
\begin{align}
    p(j)=p_{max}*\frac{N-j}{N-1},
\end{align}
where $p_{max}$ is the maximum skip probability. It can be seen that $p(j)$ will decay to 0\% at the $N$-th example. 
This guarantees that the sampling stage will not result in an empty selection. 
Once an example is selected, the subsequent selection process begins from the next example in the sequence, which means that if the $j$-th example is chosen, the next potential selection starts from the $(j+1)$-th example. 

Consequently, if the $N$-th example is selected, the Random Walk sampling terminates early, which indicates that we may fail to achieve the intended number of $n$ refined examples. Therefore, we need to make sure that this early-stop condition is scarcely activated so that the integrity of sampling can be guaranteed.
Empirically, the skipping probability should not be too large to avoid unstable training. Meanwhile, if $N$ is too large (i.e., $N \gg n$), the retrieval quality will be impaired, while if $N$ is too small (i.e., $N \approx n$), it might hurt the diversity. Therefore, as $n$ is normally less than 5 due to the context length limitation, $N$ is set to 10 as a balanced choice. For simplification, the maximum skip probability $p_{max}$ is then set to $(N-1)\%=9\%$ in this work so that $p(j)$ could be simply written as $(N-j)\%$.
In this case, let $n=2$ and $N=10$, there is only a likelihood of $\frac{(N-1)!}{100^{N-1}}=3.6288e^{-13}$ that the early-stop is activated, which is nearly indistinguishable from zero, thus generally maintaining the sampling integrity.

\subsubsection{Sequence Reversal}
Due to the training strategy and the inherent characteristic of natural languages, LLMs have difficulty capturing long-range dependencies or relationships between words that are far apart in the input text, namely the distance dependency. The positions of examples in the context might influence the generation results due to the distance dependency of LLMs~\cite{frermann-etal-2023-conflicts}. In this case, it is of significance to put the most informative example exactly near the current input query.

Formally, given the context examples, previous works like MolReGPT tend to organize the input text by directly fitting them into the context. Therefore, the context could be represented as:

\begin{align}
    \mathcal{P}(x_1,y_1)\oplus\mathcal{P}(x_2,y_2)\oplus ... \oplus \mathcal{P}(x_n,y_n),
\end{align}

where $\mathcal{P}$ denotes the prompt template and $(x_i, y_i)$ is the $i$-th refined similar molecule-caption pair, while $\oplus$ represents the concatenation.
Obviously, $(x_n,y_n)$ is generally the least informative molecule-caption pair among $n$ refined examples, while it is the closest example to the current input query. 
Here, we propose Sequence Reversal to resolve this question by simply reversing the sequence of examples. Specifically, the context can be represented as:

\begin{align}
    \mathcal{P}(x_n,y_n)\!\oplus\!\mathcal{P}(x_{n-1},y_{n-1})\!\oplus...\oplus\! \mathcal{P}(x_1,y_1).
\end{align}

\subsection{In-Context Molecule Tuning}
As shown in Figure \ref{fig:intro}, similar molecules typically share similar structures and chemical properties. Building on this principle, MolReGPT~\cite{li2023empowering} has demonstrated the effectiveness of in-context learning, which aims to prompt Large Language Models (LLMs) to learn from similar examples without the need for domain-specific pre-training and fine-tuning. However, MolReGPT heavily relies on the reasoning and in-context learning capabilities of LLMs, resulting in poor performance with relatively small language models.
To address the deficits 
, we propose In-Context Molecule Tuning to fine-tune the parameters of LLMs, enabling them to learn from context examples and reason on the current input.
Notably, In-Context Molecule Tuning can be easily adapted to any LLM, allowing even smaller language models to unlock their in-context molecule learning capability by learning the differences and similarities between molecule-caption pairs in the context. 

Formally, given the training dataset $\mathcal{D}$ and the parameters of the LLM $\theta$, let the current input of the LLM be $x$ and the target output be $y$, where $(x,y) \in D$ denotes the molecule-caption pair. Traditional supervised fine-tuning methods directly learn the mapping from the input to the output $f: x\rightarrow y$ and the loss function $\mathcal{L}^{ft}(\theta)$ could be denoted as:

\begin{align}
    \mathcal{L}^{ft}(\theta) =  \underset{(x,y)\in \mathcal{D}}{\sum}[-\log p_\theta(y|x)].
\end{align}

In contrast, ICMA does not simply conduct the next token prediction on the output part but learns the entire input in an auto-regressive manner. ICMA first employs the Hybrid Context Retrieval and Post-retrieval Re-ranking to obtain a subset $D_{(x,y)} \subset D$ containing $n$ similar examples $\{(x_i, y_i) | 1\leq i \leq n\}$, from the training set. Different from the previous ICL objective, ICMA is also motivated to learn the mapping $f_i: x_i\rightarrow y_i$ inside the context examples in an obvious manner. Notably, molecule-caption mapping is the most informative part in the context examples because similar molecule sub-structures inherit similar characteristics. For example, RCOOH (carboxyl group) usually indicates that the molecule is an acid. In this way, learning the alignment between functional groups and molecule captions in the context examples could benefit the final prediction. For simplicity, we could assume that context examples are independent of each other as the molecule-caption mapping plays the most important role in the prediction.
In this case, the aggregation of mappings $\mathcal{F}_{(x,y)}=\{f_1, f_2, ..., f_{n}\}$ could be learned from context and will altogether contribute to the final prediction with the corresponding context $C_{(x,y)}$, which wraps the context examples into the input text.  
Therefore, the objective of ICMA can be represented as:

\begin{small} 
\begin{equation} 
\mathcal{L}(\mathcal{F}_{(x,y)}) = \underset{(x_i,y_i)\in D_{(x,y)}}{\sum}-\log p_\theta(y_i|x_i) ,
\end{equation} 
\begin{equation}
    \mathcal{L}^{I\!C\!M\!A}(\theta)\!=\!\underset{(x,y)\in D}{\sum}\!\left(\!-\log p_\theta(y|x, C_{(x,y)})\!+\! \mathcal{L}(\mathcal{F}_{(x,y)})\right)\!,
\end{equation}
\end{small}

\noindent where $\mathcal{L}(\mathcal{F}_{(x,y)})$ represents the aggregated mapping loss for molecule-caption pair $(x,y)$, while $\mathcal{L}^{I\!C\!M\!A}(\theta)$ denotes the overall loss function.

By learning the context examples as well as the corresponding mappings, ICMA enables LLMs to learn the alignment between molecular and textual spaces in a more explainable manner. Moreover, ICMA could effectively harness the reasoning capabilities of LLMs and seamlessly adapt general LLMs to the task of molecule-caption translation.

\section{Experiments}
\label{sec:Experiments}
In this section, we aim to evaluate the effectiveness of ICMA. Firstly, we introduce the experimental settings.
Then, we compare ICMA with the selected baselines on the ChEBI-20 dataset and further test ICMA on a smaller dataset, PubChem324k.
Meanwhile, we also comprehensively study the factors that will affect the performance of ICMA, including retrieval algorithms, context settings, model scales, backbone models, complexity of data samples, and the Random Walk sampling strategy.
After that, an ablation study is conducted to justify the design of Post-retrieval Re-ranking components.
What's more, we compare ICMA with models that require extra-domain alignment stages to demonstrate the efficiency and effectiveness of ICMA. Furthermore, we verify ICMA on molecule property prediction tasks, illustrating the generalization capability of ICMA to other molecule-related tasks.
Finally, we conduct a case study to provide more details about our method.

\subsection{Experimental Settings}
We will first detail our experiment settings. 
All the hyper-parameters are illustrated in Table \ref{tab:hyper}. If not specifically stated, the cutoff length that truncates the inputs for LLMs to process is set to 1024, and n\_shot is set to 2 to control the variables. Notably, for LLMs with over 1 billion parameters, we apply LoRA~\cite{hu2021lora} to save the GPU memory and accelerate computation. Otherwise, we fine-tune the full model of LLMs. For the dataset, we apply two different molecule-caption translation datasets,
ChEBI-20~\cite{edwards2021text2mol} and PubChem324k~\cite{liu2023molca}. The details of the datasets are shown in Table \ref{tab:dataset}.
\begin{table}[htb]
    \centering
    \caption{Hyper-Parameters}
    \begin{tabular}{c|c}
    \toprule
    Item & Value \\
    \midrule
    batch size & 32  \\
    epochs & 10 \\
    warm-up steps & 1000 \\
    cutoff length & 512, 1024, 1536, 2048 \\
    refined examples ($n$) & 1,2,3,4 \\
    rough examples ($N$) & 10 \\
    maximum skip probability ($p_{max}$) & 9 \\
    learning rate & 2e-4 \\
    \midrule
    lora\_r & 32\\
    lora\_alpha & 64 \\
    lora\_dropout & 0.1 \\
    int8 & True \\
    fp16 & True \\
    \midrule
    temperature & 0.7 \\
    top\_p & 0.85 \\
    top\_k & 40 \\
    num\_beams & 1 \\
    max\_new\_tokens & 256 \\
    \bottomrule
    \end{tabular}
    \label{tab:hyper}
\end{table}

\begin{table}[htb]
    \centering
    \caption{Details of the datasets, ChEBI-20 and PubChem324k. For PubChem324k, we follow the split in MolCA~\cite{liu2023molca}, while ignoring the \emph{Pretrain} fold.}
    \resizebox{0.75\columnwidth}{!}{
    \begin{tabular}{c|c|c|c}
    \toprule
    Dataset & Train & Validation & Test \\
    \midrule
    ChEBI-20 &  26,407     & 3,001           & 3,000  \\
    PubChem324k & 12,000 & 1,000  & 2,000 \\
    \bottomrule
    \end{tabular}
    }
    \label{tab:dataset}
\end{table}

For comparison, we first select two different foundation LLMs as the backbones of ICMA, namely Galactica-125M~\cite{taylor2022galactica} and Mistral-7B-instruct-v0.2~\cite{jiang2023mistral}. The former is a representative and smaller LLM that has been pre-trained on unstructured scientific corpora, which have been aware of the molecular knowledge from the pre-training stage, while the latter is a general LLM with 7 billion parameters, whose capabilities are comparable to GPT-3.5-turbo. 
To study the model agnosticism, we include two more backbone LLMs (Galactica-1.3B and Llama-2-7b-chat-hf) in Section IV-F.
Notably, in the factor analysis and ablation study parts, we mainly select Galactica-125M for experiments to reduce the computational costs.
To better present our results, considering baseline models, we first select MolT5-base, MolT5-large~\cite{edwards-etal-2022-translation} and MolReGPT (GPT-3.5-turbo and GPT-4-0314)~\cite{li2023empowering} for comparison on the ChEBI-20 dataset and the case study. We also include the comparison and discussion with previous additional SOTA methods, namely BioT5~\cite{pei2023biot5}, MolXPT~\cite{liu2023molxpt}, and MolCA~\cite{liu2023molca}, in Section IV-J.

\subsection{Performance Comparison}
\begin{table*}[htb]
    \centering
    \vskip -0.15in
    \caption{Mol2Cap results on ChEBI-20 dataset (\textbf{Best}, \underline{Second Best}). Here, the results of MolT5-base and MolT5-large are domain-specific pre-training \& fine-tuning results \cite{edwards-etal-2022-translation}, while MolReGPT(GPT-3.5-turbo) and MolReGPT(GPT-4-0314) are prompting \& in-context learning results \cite{li2023empowering}. More importantly, Galactica-125M and Mistral-7B demonstrate naive supervised fine-tuned results, while ICMA(Galactica-125M) and ICMA(Mistral-7B) illustrate the ICMA results.}
    \vskip -0.05in
    \resizebox{1.8\columnwidth}{!}{
    \begin{tabular}{c|c|c|c|c|c|c}
    \toprule
    Methods & BLEU-2$\uparrow$ & BLEU-4$\uparrow$ & ROUGE-1$\uparrow$ & ROUGE-2$\uparrow$ & ROUGE-L$\uparrow$ & METEOR$\uparrow$ \\
    \midrule
    MolT5-base \cite{edwards-etal-2022-translation} & 0.540 & 0.457 & 0.634 & 0.485 & 0.578 & 0.569 \\ 
    MolReGPT (GPT-3.5-turbo) & 0.565 & 0.482 & 0.623 & 0.450 & 0.543 & 0.585 \\
    MolT5-large \cite{edwards-etal-2022-translation} & 0.594 & 0.508 & 0.654 & 0.510 & 0.594 & 0.614 \\
    MolReGPT (GPT-4-0314) & 0.607 & 0.525 & 0.634 & 0.476 & 0.562 & 0.610 \\
    \midrule
    Galactica-125M & 0.585 &	0.501 &	0.630 &	0.474 &	0.568 & 	0.591 \\
    ICMA(Galactica-125M)$_{2,2048}$ & \underline{0.636} &	\underline{0.565} 	&\underline{0.674} &	\underline{0.536} &	\underline{0.615} &	\underline{0.648} \\
    Mistral-7B &0.566 &	0.478 &	0.614 &	0.449 &	0.547 	&0.572  \\
    ICMA(Mistral-7B)$_{2,2048}$& \textbf{0.651} &	\textbf{0.581} & \textbf{0.686} & \textbf{0.550} &	\textbf{0.625} 	& \textbf{0.661} \\
    \midrule
    \bottomrule
    \end{tabular}    
    }
    \vskip -0.15in
    \label{tab:m2c}
\end{table*}

\begin{table*}[htb]
    \centering      
    \caption{Cap2Mol results on ChEBI-20 dataset (\textbf{Best}, \underline{Second Best}). Here, the results of MolT5-base and MolT5-large are domain-specific pre-training \& fine-tuning results \cite{edwards-etal-2022-translation}, while MolReGPT(GPT-3.5-turbo) and MolReGPT(GPT-4-0314) are prompting \& in-context learning results \cite{li2023empowering}. More importantly, Galactica-125M and Mistral-7B demonstrate naive supervised fine-tuned results, while ICMA(Galactica-125M) and ICMA(Mistral-7B) illustrate the ICMA results.}
    \vskip -0.05in
    \resizebox{1.8\columnwidth}{!}{
    \begin{tabular}{c|c|c|c|c|c|c|c}
    \toprule
    Method & BLEU$\uparrow$ & EM$\uparrow$ & Levenshtein$\downarrow$ & MACCS FTS$\uparrow$ & RDK FTS$\uparrow$ & Morgan FTS$\uparrow$ & Validity$\uparrow$ \\
    \midrule
    MolT5-base \cite{edwards-etal-2022-translation} & 0.769 & 0.081 & 24.458 & 0.721 & 0.588 & 0.529 & 0.772 \\
    MolReGPT(GPT-3.5-turbo) & 0.790 & 0.139 & 24.91 & 0.847 & 0.708 & 0.624 & 0.887 \\ 
    MolT5-large \cite{edwards-etal-2022-translation} & 0.854 & 0.311 & \textbf{16.071} & 0.834 & 0.746 & 0.684 & 0.905 \\
    MolReGPT(GPT-4-0314) & \textbf{0.857} & 0.280 & \underline{17.14} & \underline{0.903} & 0.805 & 0.739 & 0.899 \\
    \midrule
    Galactica-125M & 0.781 & 	0.173 &	26.34 &	0.836 &	0.708 &	0.631 &	0.916  \\
    ICMA(Galactica-125M)$_{4,2048}$ & 0.843 &	\underline{0.391} &	17.71 &	0.897 &	\underline{0.812}& 	\underline{0.753} &	\underline{0.941}   \\
    Mistral-7B & 0.767 &	0.234 &	27.39 &	0.852 &	0.718 &	0.649 &	0.918   \\
    ICMA(Mistral-7B)$_{4,2048}$&  \underline{0.855} & \textbf{0.460} & 18.73 & \textbf{0.916} & \textbf{0.837} & \textbf{0.789} & \textbf{0.958}\\
    \bottomrule
    \end{tabular}
    }
    \vskip -0.15in
    \label{tab:c2m}
\end{table*}
We compare and analyse the performance of ICMA with previous baselines and their original foundation models with naive supervised fine-tuning (SFT) from the two subtasks of molecule caption translation, namely Mol2Cap and Cap2Mol.

\noindent\textbf{Mol2Cap}. As illustrated in Table \ref{tab:m2c}, Galactica-125M with SFT has already shown competitive performance to previous baselines due to its pre-training on scientific corpora. However, ICMA could still improve the performance of Galactica-125M by 12.8\% and 8.3\% considering the BLEU-4 and ROUGE-L scores on the ChEBI-20 dataset. With only 125 million parameters, ICMA(Galactica-125M) can beat MolT5-large, which owns more than 780 million parameters.
Meanwhile, the general LLM, Mistral-7B with naive SFT, only achieves a performance that is slightly better than MolT5-base, despite the fact that Mistral-7B is 70 times larger than MolT5-base. 
This outcome is not surprising because Mistral-7B is not specifically designed or pre-trained for biomolecular purposes.
It also reveals that although general LLMs have illustrated powerful capabilities with billions of parameters, few works have adapted them to the biomolecular domain due to their unsatisfactory fine-tuning performance. 
However, with ICMA, Mistral-7B could easily demonstrate its superior in-context learning and reasoning capabilities and perform its advantage of parameters.
As a result, ICMA(Mistral-7B) achieves the best performance across all the models, obtaining 0.581 BLEU-4 and 0.661 METEOR scores on the ChEBI-20 dataset, which is even better than the domain-specific pre-trained Galactica-125M.

\noindent\textbf{Cap2Mol}. Similarly, as shown in Table \ref{tab:c2m}, compared to their original foundation models with SFT, ICMA significantly boosts the molecule generation performance. Notably, ICMA(Mistral-7B) achieves state-of-the-art molecule generation performance, generating 46.0\% exactly matched molecules, which nearly doubles the results of naive supervised fine-tuned Mistral-7B. Although both ICMA(Galactica-125M) and ICMA(Mistral-7B) achieve higher Levenshtein scores, due to the characteristic of the metric, a lower Levenshtein score does not mean better generation quality. For example, the Levenshtein score of $CO \rightarrow CCOC$ is 2, and the Levenshtein score of $CCC \rightarrow CCOC$ is just 1. However, the former molecule is considered more similar to the target molecule as they share the same functional group.
Instead, if we look at the molecule fingerprints similarity scores, ICMA(Mistral-7B) could achieve superior results and obtain the best validity score, showing better molecule generation quality.

Additionally, experiments are also conducted on a smaller dataset, PubChem324k. As shown in Table \ref{tab:m2c_pub} and \ref{tab:c2m_pub}, ICMA could still boost the Mol2Cap performance of LLMs like Galactica-125M and Mistral-7B, which further proves the generalization of ICMA in the molecule-caption translation task.

\begin{table}[htbp]
    \centering
    \caption{Mol2Cap results on PubChem324k dataset (\textbf{Best}, \underline{Second Best}). Here, Galactica-125M and Mistral-7B demonstrate naive supervised fine-tuned results, while ICMA(Galactica-125M) and ICMA(Mistral-7B) illustrate the ICMA results.}
    \resizebox{1.0\columnwidth}{!}{
    \begin{tabular}{c|c|c|c|c|c|c}
    \toprule
    Method & BLEU-2$\uparrow$ & BLEU-4$\uparrow$ & ROUGE-1$\uparrow$ & ROUGE-2$\uparrow$ & ROUGE-L$\uparrow$ & METEOR$\uparrow$ \\
    \midrule
    Galactica-125M & 0.333	& 0.265&	0.465&	0.322&	0.417	&0.406\\
    ICMA(Galactica-125M)$_{4,2048}$& \underline{0.411}	&\underline{0.338}&	\underline{0.497} &\underline{0.359}	&\underline{0.446}&	\underline{0.457}\\
    Mistral-7B & 0.361&	0.288	&0.471&	0.325&	0.419&	0.421   \\
    ICMA(Mistral-7B)$_{4,2048}$& \textbf{0.416}&	\textbf{0.345}&	\textbf{0.505}&	\textbf{0.367}&	\textbf{0.453}&	\textbf{0.464}\\
    \bottomrule
    \end{tabular}
    }

    \label{tab:m2c_pub}
\end{table}

\begin{table}[htbp]
    \centering
    \caption{Cap2Mol results on PubChem324k dataset (\textbf{Best}, \underline{Second Best}). Here, Galactica-125M and Mistral-7B demonstrate naive supervised fine-tuned results, while ICMA(Galactica-125M) and ICMA(Mistral-7B) illustrate the ICMA results.}
    \resizebox{1.0\columnwidth}{!}{
    \begin{tabular}{c|c|c|c|c|c|c|c}
    \toprule
    Method & BLEU$\uparrow$ & EM$\uparrow$ & Levenshtein$\downarrow$ & MACCS FTS$\uparrow$ & RDK FTS$\uparrow$ & Morgan FTS$\uparrow$ & Validity$\uparrow$ \\
    \midrule
    Galactica-125M &0.485&	0.031	&\underline{62.08}&	0.681&	0.510	&0.403	& 0.835\\
    ICMA(Galactica-125M)$_{4,2048}$& \textbf{0.600}	& \underline{0.098}	&\textbf{49.00}	&\underline{0.764}	&\underline{0.632}	&\underline{0.523}	&\underline{0.909}\\
    Mistral-7B & 0.438&	0.082	&74.16&	0.731&	0.577&	0.472	&0.866   \\
    ICMA(Mistral-7B)$_{4,2048}$& \underline{0.526} &	\textbf{0.163}	&62.25	&\textbf{0.799}	&\textbf{0.678}	&\textbf{0.573}	&\textbf{0.935}\\
    \bottomrule
    \end{tabular}
    }
    
    \label{tab:c2m_pub}
\end{table}

\subsection{Study of Retrieval Algorithms}
\begin{table}[htbp]
    \centering
    \caption{Performance comparison of different retrieval algorithms for Mol2Cap task (\textbf{Best}, \underline{Second Best}). The backbone is ICMA(Galactica-125M)$_{2,1024}$.}
    \resizebox{1.0\columnwidth}{!}{
    \begin{tabular}{c|c|c|c|c|c|c}
    \toprule
    Method & BLEU-2$\uparrow$ & BLEU-4$\uparrow$ & ROUGE-1$\uparrow$ & ROUGE-2$\uparrow$ & ROUGE-L$\uparrow$ & METEOR$\uparrow$ \\
    \midrule
    Random & 0.611 &	0.533 &	0.648 &	0.500 &	0.587 &	0.613 \\ 
    Morgan FTS & \underline{0.627} &	\underline{0.552} &	\underline{0.661} &	\underline{0.517} &	\underline{0.600} &	\underline{0.633}\\
    Mole-BERT & \textbf{0.641} &	\textbf{0.568} &	\textbf{0.671} &	\textbf{0.531} &	\textbf{0.611} &	\textbf{0.645}\\
    \bottomrule
    \end{tabular}
    }
    \vskip -0.2in
    \label{tab:m2c_algo}
\end{table}

\begin{table}[htbp]
    \centering
    \caption{Performance comparison of different retrieval algorithms for Cap2Mol task (\textbf{Best}, \underline{Second Best}). The backbone is ICMA(Galactica-125M)$_{2,1024}$.}
    \resizebox{1.0\columnwidth}{!}{
    \begin{tabular}{c|c|c|c|c|c|c|c}
    \toprule
    Method & BLEU$\uparrow$ & EM$\uparrow$ & Levenshtein$\downarrow$ & MACCS FTS$\uparrow$ & RDK FTS$\uparrow$ & Morgan FTS$\uparrow$ & Validity$\uparrow$ \\
    \midrule
    Random &  0.823 &	0.299 &	20.24 &	0.873 &	0.772 &	0.709 &	0.928 \\
    SBERT & \underline{0.828} &	\underline{0.322} &	\underline{19.91} &	\underline{0.880} &	\underline{0.786} &	\underline{0.720} &	\underline{0.929}\\
    BM25 & \textbf{0.842} &	\textbf{0.355} &	\textbf{17.99} &	\textbf{0.890} &	\textbf{0.805} &	\textbf{0.741} &	\textbf{0.935}\\
    \bottomrule
    \end{tabular}
    }
    
    \label{tab:c2m_algo}
\end{table}
We also study the influence of retrieval algorithms to illustrate the importance of retrieval qualities. For molecule retrieval, we compare the new proposed Molecule Graph Retrieval using Mole-BERT with random retrieval and the Morgan FTS retrieval. As shown in Table \ref{tab:m2c_algo}, Mole-BERT achieves the best results on all of the metrics, proving its superiority in molecule retrieval.
For the caption retrieval, we compare BM25 Caption Retrieval with random retrieval and SBERT retrieval under the framework of ICMA.
As depicted in Table \ref{tab:c2m_algo}, BM25 Caption Retrieval illustrates its excellent performance among the three caption retrieval methods.

\subsection{Study of Context Settings}
\begin{figure}
    \centering
    \includegraphics[width=1.0\columnwidth]{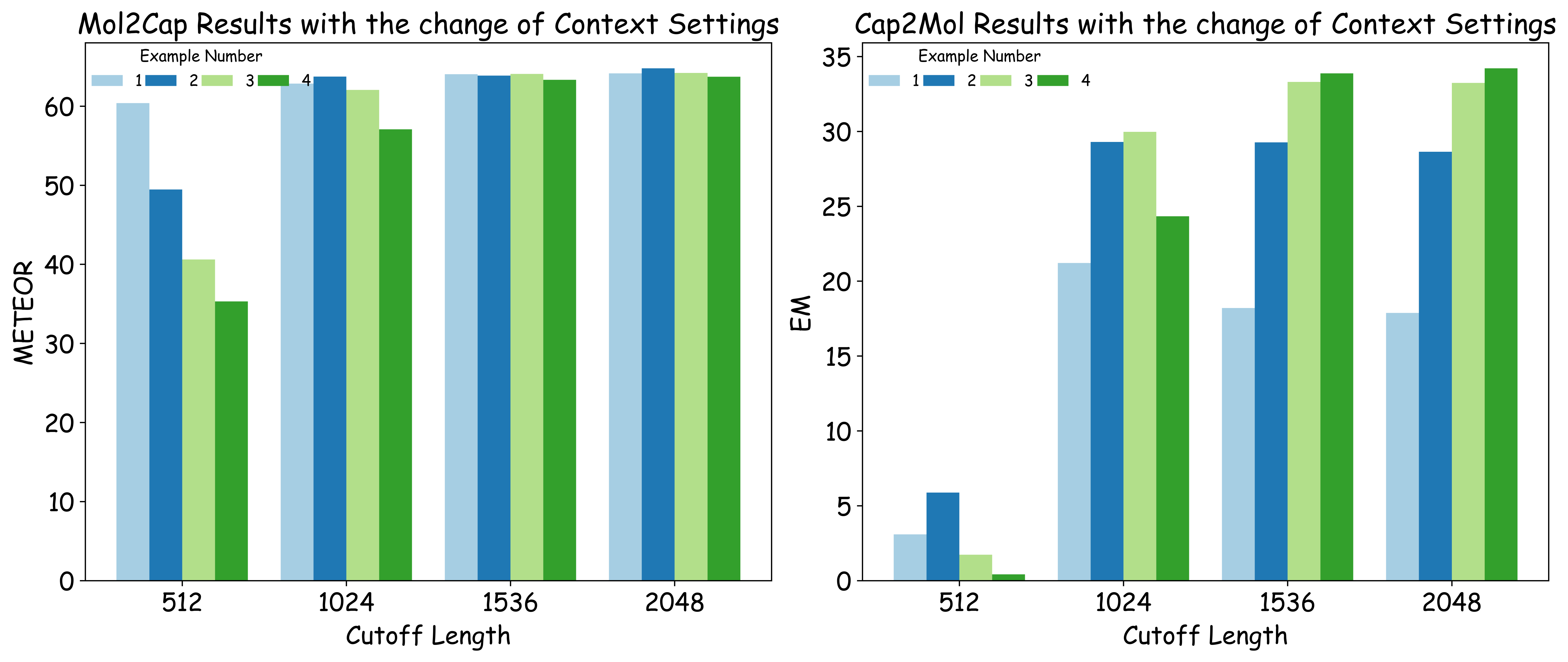}
    \caption{The model performance with the change of context settings, including the number of refined examples (i.e., $n$) and cutoff length. Mol2Cap Results (Left) and Cap2Mol Results (Right).}
    \label{fig:model_length}
\end{figure}
In ICMA, the context settings, including the example number and cutoff length, are also important to its performance. The increase in example number will also drastically require longer context length. If the context length is longer than the cutoff length, then the training will be insufficient because most of the inputs are cropped, and the information is lost. Meanwhile, during the inference stage, the context length also influences the information that LLMs can take.
In this case, we want to make sure that most of the context examples fit in the cutoff length. Considering that the input length limitation of Galactica-125M is 2048, and the model series has no length extrapolating capability, we test the cutoff length within the range of \{512, 1024, 1536, 2048\} and the example number from 1 to 4 for analysis.
As illustrated in Figure \ref{fig:model_length}, when the example number increases, the performance becomes worse if the cutoff length is too short. Notably, when the cutoff length is set to 512, the context is not complete for LLMs to consistently learn valuable knowledge from it, leading to worse results. However, when the cutoff length is long enough, the performance mainly becomes better as the increased example number provides more information to help LLMs make accurate predictions. This phenomenon is more obvious in the Cap2Mol task, while in the Cap2Mol task, the impact of context settings is not significant, as there is a balance between the extra information and noises with the increase of cutoff length and context example numbers.

\subsection{Study of Scaling Law}
\begin{figure}
    \centering
    \includegraphics[width=\columnwidth]{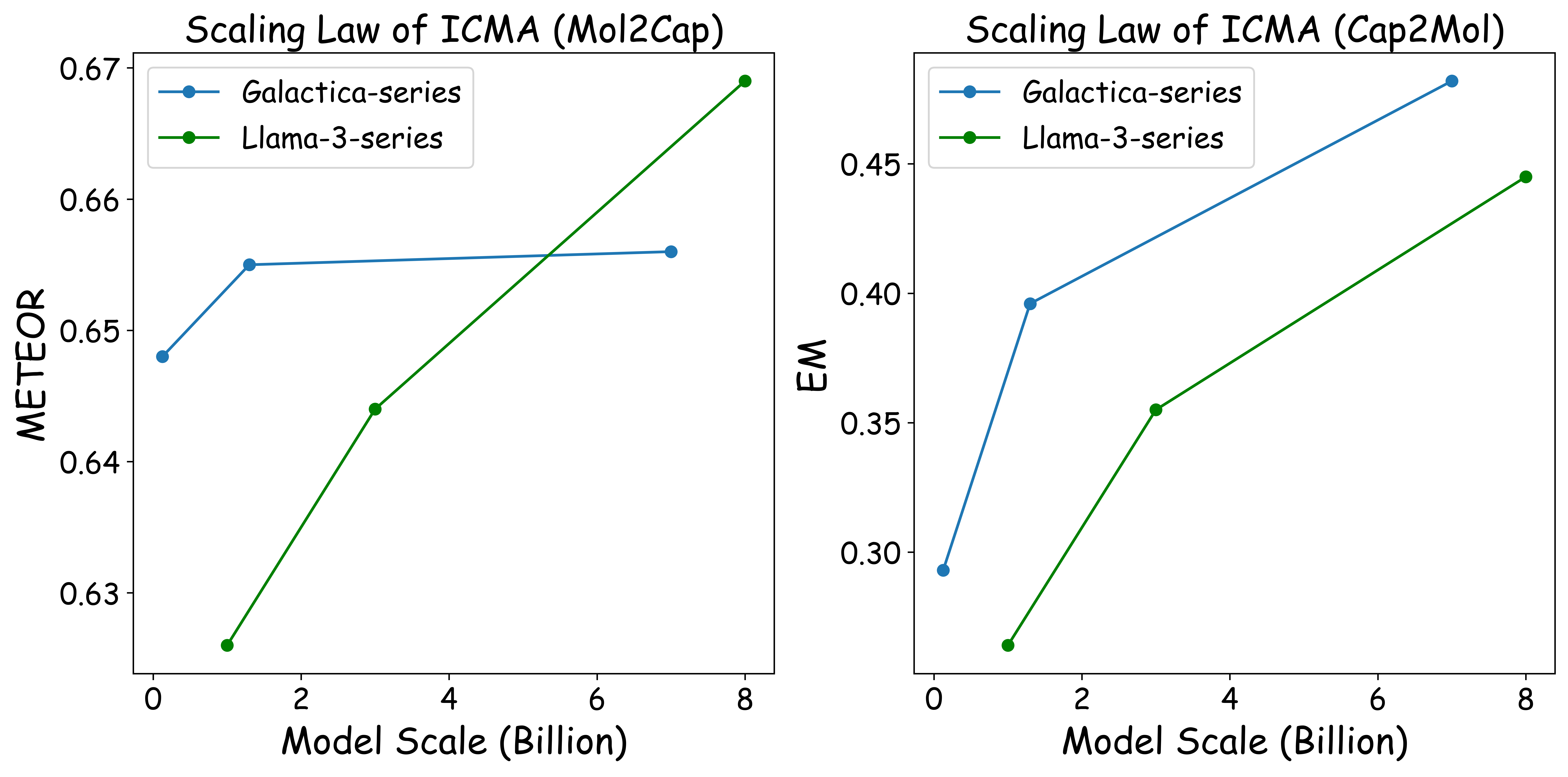}
    \caption{The scaling law of ICMA. Three models with different levels of parameters are selected, including the Galactica series (Blue) and the Llama-3 series (Green). Mol2Cap Results (Left) and Cap2Mol Results (Right).}
    \label{fig:model_scale}
\end{figure}
As ICMA creates a new paradigm for adapting powerful LLMs with billion parameters to the molecule-caption translation task, it is interesting to study the relationship between the performance and model scales. In this case, we select LLMs with different scales of parameters, ranging from 125 million to 8 billion, to study the scaling law of ICMA. Specifically, we select the Galactica series and Llama-3 series for demonstration.

As illustrated in Figure \ref{fig:model_scale}, with the increase of model scale, the performance of ICMA for the Llama-3 series improves in both the Mol2Cap and Cap2Mol tasks. The improvements in molecule understanding and generation capabilities significantly benefit from the increased model scales. In the case of the Galactica series, when the model size is increased from 1.3 billion to 6.7 billion parameters, a consistent pattern is observed in the Cap2Mol task. However, the performance on the Mol2Cap task exhibits a notably different trend, with minimal improvement observed. This discrepancy could indicate underlying deficiencies in the model’s natural language understanding and generation capabilities, which caps Galactica‘s performance in the Mol2Cap task.

\subsection{Study of Model Agnosticism}
To highlight the model agnosticism of ICMA, we also expand our experiments on two more backbone LLMs, Galactica-1.3B and Meta-Llama-3-8B-Instruct. The results are shown in Tables \ref{tab:m2c_more} and \ref{tab:c2m_more}. It is evident that ICMA enhances the performance of both backbone models, which further proves the versatility and model agnosticism of ICMA.
\begin{table}[htbp]
    \centering
    \caption{Performance comparison of two more backbone LLMs (Galactica-1.3B and Meta-Llama-3-8B-Instruct) for Mol2Cap task on ChEBI-20 dataset (\textbf{Best}, \underline{Second Best}).}
    \resizebox{1.0\columnwidth}{!}{
    \begin{tabular}{c|c|c|c|c|c|c}
    \toprule
    Method & BLEU-2$\uparrow$ & BLEU-4$\uparrow$ & ROUGE-1$\uparrow$ & ROUGE-2$\uparrow$ & ROUGE-L$\uparrow$ & METEOR$\uparrow$ \\
    \midrule
    Galactica-1.3B & 0.589 &	0.507 &	0.638 &	0.484 &	0.574 &	0.597 \\ 
    ICMA(Galactica-1.3B) & 0.602 &	0.534 &	\underline{0.668} &	\underline{0.530} &	\underline{0.607} &	\underline{0.649} \\ 
    Llama3-8B & \underline{0.618} &	\underline{0.542} &	0.660 &	0.515 &	0.599 &	0.623\\
    ICMA(Llama3-8B) & \textbf{0.665} &	\textbf{0.595} &	\textbf{0.693} &	\textbf{0.559} &	\textbf{0.633} &	\textbf{0.669}\\
    \bottomrule
    \end{tabular}
    }
    \label{tab:m2c_more}
    \vskip -0.2in
\end{table}

\begin{table}[htbp]
    \centering
    \caption{Performance comparison of two more backbone LLMs (Galactica-1.3B and Meta-Llama-3-8B-Instruct) for Cap2Mol task on ChEBI-20 dataset (\textbf{Best}, \underline{Second Best}).}
    \resizebox{1.0\columnwidth}{!}{
    \begin{tabular}{c|c|c|c|c|c|c|c}
    \toprule
    Method & BLEU$\uparrow$ & EM$\uparrow$ & Levenshtein$\downarrow$ & MACCS FTS$\uparrow$ & RDK FTS$\uparrow$ & Morgan FTS$\uparrow$ & Validity$\uparrow$ \\
    \midrule
    Galactica-1.3B & 0.812 & 0.292 &	22.47 &	0.882 &	0.777 &	0.709 &	\underline{0.950} \\
    ICMA(Galactica-1.3B) &  \underline{0.839} &	\underline{0.396} &	21.61&	\underline{0.910} &	\underline{0.829} &	\underline{0.770} &	0.946 \\
    Llama3-8B & 0.826 &	0.356 &	\underline{20.70} &	0.894 &	0.793 &	0.733 &	0.939\\
    ICMA(Llama3-8B) & \textbf{0.851} &	\textbf{0.445} &\textbf{19.27} &	\textbf{0.915} &	\textbf{0.836} &	\textbf{0.785} &	\textbf{0.958} \\
    \bottomrule
    \end{tabular}
    }
    \label{tab:c2m_more}
\end{table}

\subsection{Performance for Complex Molecules and Captions}
ICMA also works better for complex molecular structures and molecule captions. Normally, the lengthy molecules can be more complex for LLMs. 
In this case, 
we extracted a subset of 747 complex molecules, whose SMILES strings have more than 100 characters and a subset of 221 complex captions with more than 500 characters from the ChEBI-20 test set. We compared the performance of ICMA with previous baselines on this subset. Our experimental results shown in Table \ref{tab:mol2cap_complex} and \ref{tab:cap2mol_complex} indicate that ICMA outperforms the previous baselines, especially in the Cap2Mol task, as similar examples learned from ICMT typically share comparable levels of complexity, which helps better aligns molecules with texts. 

\begin{table}[htbp]
    \centering
    \caption{The Mol2Cap performance comparison on the subset of ChEBI-20 with complex molecules whose SMILES strings have more than 100 characters (\textbf{Best}, \underline{Second Best}).}
    \resizebox{1.0\columnwidth}{!}{
    \begin{tabular}{c|c|c|c|c|c|c}
    \toprule
        Method & BLEU-2$\uparrow$ & BLEU-4$\uparrow$ & ROUGE-1$\uparrow$ & ROUGE-2$\uparrow$ & ROUGE-L$\uparrow$ & METEOR$\uparrow$ \\ \midrule
        MolT5-large & \underline{0.684} & \underline{0.624} & \underline{0.715} & \underline{0.590} & \underline{0.661} & \underline{0.685} \\
        MolReGPT(GPT-4-0314) & 0.670 & 0.603 & 0.692 & 0.555 & 0.630 & 0.673 \\
        ICMA(Mistral-7B) & \textbf{0.687} & \textbf{0.632} & \textbf{0.727} & \textbf{0.611} & \textbf{0.673} & \textbf{0.713} \\ 
    \bottomrule
    \end{tabular}
    }
    \label{tab:mol2cap_complex}
\end{table}

\begin{table}[htbp]
    \centering
    \caption{The Cap2Mol performance comparison on the subset of ChEBI-20 with complex captions that have more than 500 characters (\textbf{Best}, \underline{Second Best}).}
    \resizebox{1.0\columnwidth}{!}{
    \begin{tabular}{c|c|c|c|c|c|c|c}
    \toprule
        Method & BLEU$\uparrow$ & EM$\uparrow$ & Levenshtein$\downarrow$ & MACCS FTS$\uparrow$ & RDK FTS$\uparrow$ & Morgan FTS$\uparrow$ & Validity$\uparrow$ \\ \midrule
        MolT5-large & 0.624 & 0.113 & \underline{50.71} & 0.801 & 0.669 & 0.584 & \underline{0.792} \\ 
        MolReGPT(GPT-4-0314) & \textbf{0.794} & \underline{0.122} & \textbf{40.00} & \underline{0.858} & \underline{0.704} & \underline{0.630} & 0.769 \\ 
        ICMA(Mistral-7B) & \underline{0.684} & \textbf{0.235} & 70.79 & \textbf{0.883} & \textbf{0.736} & \textbf{0.670} & \textbf{0.873} \\ 
    \bottomrule
    \end{tabular}
    }
    \label{tab:cap2mol_complex}
\end{table}

\subsection{Random Walk Stability \& Maximum Skip Probability}

Due to the randomness we introduced during the Random Walk selection, there might be concerns about the stability of the strategy. However, considering that the design of the skip probability is to limit it within the maximum skip probability, which is a rather low value, and gradually decays the likelihood to ensure that at least one example is selected, we manage the randomness brought by Random Walk selection within an acceptable range. To verify this, we implemented a bucket sampling strategy for comparison. We randomly chose three different random seeds and calculated the mean and variance of the performance results for three runs. The results are demonstrated in Table \ref{tab:mol2cap_stab} and \ref{tab:cap2mol_stab}. 

\begin{table}[htbp]
    \centering
    \caption{Stability Comparison on the Mol2Cap task between two re-ranking strategy, Random Walk and Bucket Sampling. The results are with significant figures for precision (\textbf{Best}).}
    \resizebox{1.0\columnwidth}{!}{
    \begin{tabular}{c|c|c|c|c|c|c}
    \toprule
        Method & BLEU-2$\uparrow$ & BLEU-4$\uparrow$ & ROUGE-1$\uparrow$ & ROUGE-2$\uparrow$ & ROUGE-L$\uparrow$ & METEOR$\uparrow$ \\ \midrule
        Random Walk & \textbf{0.6392±0.0032} & \textbf{0.5658±0.0031} & \textbf{0.6703±0.0028} & \textbf{0.5306±0.0026} & \textbf{0.6107±0.0030} & \textbf{0.6437±0.0031} \\
        Bucket Sampling & 0.6364±0.0035 & 0.5636±0.0040 & 0.6687±0.0032 & 0.5293±0.0033 & 0.6087±0.0026 & 0.6424±0.0043 \\ 
    \bottomrule
    \end{tabular}
    }
    \label{tab:mol2cap_stab}
\end{table}

\begin{table}[htbp]
    \centering
    \caption{Stability Comparison on the Cap2Mol task between two re-ranking strategy, Random Walk and Bucket Sampling. The results are with four significant figures for precision (\textbf{Best}). }
    \resizebox{1.0\columnwidth}{!}{
    \begin{tabular}{c|c|c|c|c|c|c|c}
    \toprule
        Method & BLEU$\uparrow$ & EM$\uparrow$ & Levenshtein$\downarrow$ & MACCS FTS$\uparrow$ & RDK FTS$\uparrow$ & Morgan FTS$\uparrow$ & Validity$\uparrow$ \\ \midrule
        Random Walk & \textbf{0.8403±0.0024} & \textbf{0.3533±0.0026} & \textbf{18.45±0.56} & \textbf{0.8907±0.0020} & \textbf{0.8042±0.0033} & \textbf{0.7412±0.0031} & 0.9336±0.0027 \\ 
        Bucket Sampling & 0.8371±0.0014 & 0.3464±0.0063 & 18.58±0.23 & 0.8885±0.0020 & 0.8000±0.0021 & 0.7367±0.0021 & \textbf{0.9347±0.0056} \\ 
    \bottomrule
    \end{tabular}
    }
    \label{tab:cap2mol_stab}
\end{table}

From the results, we can observe that although the variance of the Random Walk is larger than that of bucket sampling on metrics like BLEU,  Levenshtein, RDK FTS, and Morgan FTS in the Cap2Mol task and ROUGE-L in the Mol2Cap task, the Random Walk strategy still achieves higher means, and the variances are still manageable. On some metrics, the variances of the Random Walk are even lower than that of bucket sampling, indicating that despite the inherent randomness, our method overall remains stable within normal fluctuation ranges.

\begin{figure}[htbp]
    \centering
    \includegraphics[width=\columnwidth]{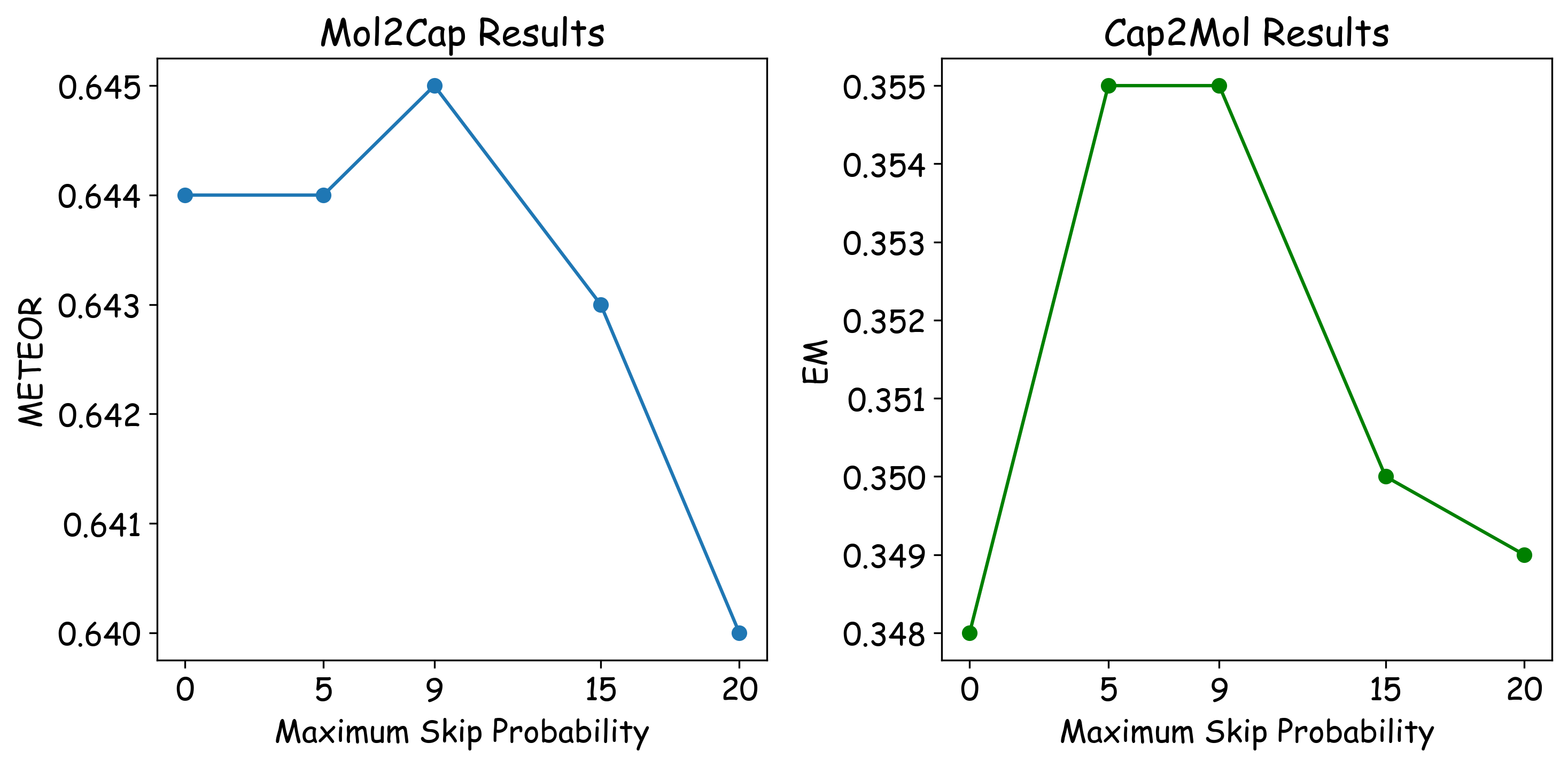}
    \caption{The performance of Galatcica-125M with the maximum skip probability $p_{max}$. Mol2Cap Results (Left) and Cap2Mol Results (Right).}
    \label{fig:skip_prob}
    \vskip -0.1in
\end{figure}

What's more, we further conduct a hyperparamter search to help decide the best maximum skip probability $p_{max}$ under the setting of $n=2$, $N=10$, and  $cuto\!f\!f\_len = 1024$. As shown in Figure \ref{fig:skip_prob}, we could obviously see that if the maximum skip probability is too large, the performance naturally drops because it is more likely to select less similar examples. However, with the assistance of skip probability decay as well as the early-stop, the performance gap is still marginal. Among all the selected maximum probabilities, $p_{max}=9\%$ achieves the highest exact match score in the Cap2Mol task and the best METEOR score in the Mol2Cap task.
However, we must acknowledge that the Random Walk strategy inherently introduces a degree of randomness, and for different numbers of rough examples $N$ and refined examples $n$, the best maximum probability might vary, which means the observed results may not fully reflect the true potential of the Random Walk sampling.

\subsection{Ablation Study}
\begin{table}[htbp]
    \centering
    \caption{Ablating components of Post-retrieval Re-ranking for Mol2Cap task (\textbf{Best}, \underline{Second Best}). The backbone is ICMA(Galactica-125M)$_{2,1024}$.}
    \resizebox{1.0\columnwidth}{!}{
    \begin{tabular}{c|c|c|c|c|c|c}
    \toprule
    Method & BLEU-2$\uparrow$ & BLEU-4$\uparrow$ & ROUGE-1$\uparrow$ & ROUGE-2$\uparrow$ & ROUGE-L$\uparrow$ & METEOR$\uparrow$ \\
    \midrule
    ICMA & \textbf{0.641} &	\textbf{0.568} &	\textbf{0.671} &	\textbf{0.531} &	\textbf{0.611} &	\textbf{0.645}\\
    w/o Random Walk &0.638(3)& 	\underline{0.565(5)} &	\underline{0.670} &	0.530(3) &	\underline{0.610(3)} &	0.643(7)\\
    w/o Sequence Reverse& \underline{0.638(4)} &	0.566(3) 	&0.669 &	\underline{0.530(5)} 	&0.610(1) &	\underline{0.644(1)}\\
    \bottomrule
    \end{tabular}
    }
    \vskip -0.2in
    \label{tab:m2c_ablation}
\end{table}

\begin{table}[htbp]
    \centering
    \caption{Ablating components of Post-retrieval Re-ranking for Cap2Mol task (\textbf{Best}, \underline{Second Best}). The backbone is ICMA(Galactica-125M)$_{2,1024}$.}
    \resizebox{1.0\columnwidth}{!}{
    \begin{tabular}{c|c|c|c|c|c|c|c}
    \toprule
    Method & BLEU$\uparrow$ & EM$\uparrow$ & Levenshtein$\downarrow$ & MACCS FTS$\uparrow$ & RDK FTS$\uparrow$ & Morgan FTS$\uparrow$ & Validity$\uparrow$ \\
    \midrule
    ICMA & \textbf{0.842} &	\textbf{0.355} &	\textbf{17.99} 	&\textbf{0.890} 	&\textbf{0.805}	&\textbf{0.741} 	& \underline{0.935} \\
    w/o Random Walk & \underline{0.840} &	0.348 &	\underline{18.01} &	0.888 	&\underline{0.804} &	0.738 &	\textbf{0.939} \\
    w/o Sequence Reverse & 0.835 &	\underline{0.354} &	18.49 &	\underline{0.889} 	& 0.800 &	\underline{0.740} &	0.933 \\
    \bottomrule
    \end{tabular}
    }
    
    \label{tab:c2m_ablation}
\end{table}
Ablation study is also conducted to illustrate how the Post-retrieval Re-ranking components affect the predictions. 
We conduct ablation experiments for the Mol2Cap and Cap2Mol sub-tasks by deactivating the Random Walk and Sequence Reversal.
As shown in Table \ref{tab:m2c_ablation} and \ref{tab:c2m_ablation}, without Random Walk or Sequence Reversal, the performance of ICMA drops, though not significantly but consistently, in both tasks, proving the effectiveness of Post-retrieval Re-ranking.

\subsection{Comparative Analysis of Models with Extra Domain Alignment Stages}
In our pursuit to comprehensively evaluate the efficacy of ICMA, we have extended our comparative analysis to include models that adopt extra domain alignment stages, such as continual pre-training on chemical corpora (BioT5~\cite{pei2023biot5} \& MolXPT~\cite{liu2023molxpt}) and graph modality alignment (MolCA ~\cite{liu2023molca}. The results, as detailed in Table \ref{tab:m2c_morebase}, indicate that ICMA outperforms all the models trained with extra domain alignment in terms of BLEU-2, BLEU-4, and METEOR scores, while also achieving the second-highest ROUGE scores. Notably, the performance is achieved without introducing extra domain alignment stages, which further proves the superiority of ICMA. 

Notably, compared to MolCA, a multi-modal method that leverages 2D molecular graphs to augment LLMs for the Mol2Cap task, our ICMA does not require extra graph information and modality alignment training. Meanwhile, ICMA is capable of refining the alignment between molecular and textual representations not only for Mol2Cap task but also for Cap2Mol task, showing better versatility in aligning molecular space with textual space.

On the other hand, for the \textbf{Cap2Mol} task, we have included BioT5 and MolXPT in our comparative analysis. As presented in Table \ref{tab:c2m_morebase}, ICMA achieves the highest scores in exact match and molecule fingerprints metrics, while the validity of molecule generation is just slightly below the BioT5 and MolXPT due to the deficiency in molecular knowledge, indicating competitive performance in molecule generation. Furthermore, it is also important to note that ICMA does not require pre-training LLMs on a large-scale chemical corpus, which makes ICMA a better choice in low-data scenarios.

\begin{table}[htbp]
    \centering
    \caption{Performance comparison with models trained with extra domain alignment stages for Mol2Cap task on ChEBI-20 dataset. Here, we select BioT5~\cite{pei2023biot5}, MolXPT~\cite{liu2023molxpt}, and MolCA~\cite{liu2023molca} for comparison (\textbf{Best}, \underline{Second Best}).}
    \resizebox{1.0\columnwidth}{!}{
    \begin{tabular}{c|c|c|c|c|c|c}
    \toprule
    Backbone & BLEU-2$\uparrow$ & BLEU-4$\uparrow$ & ROUGE-1$\uparrow$ & ROUGE-2$\uparrow$ & ROUGE-L$\uparrow$ & METEOR$\uparrow$ \\
    \midrule
    BioT5~\cite{pei2023biot5} & \underline{0.635} &	\underline{0.556} &	\textbf{0.692} &	\textbf{0.559} &	\textbf{0.633} &	\underline{0.656} \\ 
    MolXPT~\cite{liu2023molxpt} & 0.594 &	0.505 &	0.660 &	0.511 &	0.597 &	0.626 \\ 
    MolCA~\cite{liu2023molca} & 0.620 &	0.531 &	0.681 &	0.537 &	0.618 &	0.651\\
    ICMA(Mistral-7B) & \textbf{0.651} &	\textbf{0.581} & \underline{0.686} & \underline{0.550} &	\underline{0.625} 	& \textbf{0.661} \\
    \bottomrule
    \end{tabular}
    }
    
    \label{tab:m2c_morebase}
\end{table}

\begin{table}[htbp]
    \centering
    \caption{Performance comparison with models trained with extra domain alignment stages for Cap2Mol task on ChEBI-20 dataset. Here, we include the results from BioT5~\cite{pei2023biot5} and MolXPT~\cite{liu2023molxpt} for comparison (\textbf{Best}, \underline{Second Best}).}
    \resizebox{1.0\columnwidth}{!}{
    \begin{tabular}{c|c|c|c|c|c|c|c}
    \toprule
    Backbone & BLEU$\uparrow$ & EM$\uparrow$ & Levenshtein$\downarrow$ & MACCS FTS$\uparrow$ & RDK FTS$\uparrow$ & Morgan FTS$\uparrow$ & Validity$\uparrow$ \\
    \midrule
    BioT5~\cite{pei2023biot5}& \textbf{0.867} & \underline{0.413} &	\textbf{15.10} &	\underline{0.886} &	\underline{0.801} &	\underline{0.734} & \textbf{1.000} \\
    MolXPT~\cite{liu2023molxpt}& - & 0.215 &	22.47 &	0.859 &	0.757 &	0.667 &	\underline{0.983} \\
    ICMA(Mistral-7B) & \underline{0.855} & \textbf{0.460} & 18.73 & \textbf{0.916} & \textbf{0.837} & \textbf{0.789} & 0.958\\
    \bottomrule
    \end{tabular}
    }
    
    \label{tab:c2m_morebase}
\end{table}

\subsection{ICMA for Molecule Property Prediction}
Although ICMA is targeted at the task of molecule-caption translation, we are curious about whether ICMA could also benefit a wide range of molecule-related tasks, such as molecule property prediction. Therefore, we conduct experiments on the MoleculeNet dataset \cite{wu2018moleculenet} and mainly focus on classification subtasks like BACE, BBBP, HIV, TOX21, and SIDER. The results are shown in Table \ref{tab:molenet}. To our surprise, ICMA shows great generalization capability to these molecule property prediction tasks, increasing the performance of both Galactica-125M and Mistral-7B. More significantly, ICMA enables the general LLM, Mistral-7B, to achieve an average improvement of 33.26\% on the five subtasks, which further proves that LLMs are inherently in-context molecule learners.
\begin{table}[htbp]
    \centering
    \caption{Molecule Property Prediction Performance of ICMA compared with directly fine-tuning the backbone models on the MoleculeNet Dataset (\textbf{Best}, \underline{Second Best}).}
    \resizebox{1.0\columnwidth}{!}{
    
    \begin{tabular}{c|c|c|c|c|c}
    \toprule
        Method & BACE$\uparrow$ & BBBP$\uparrow$ & HIV$\uparrow$ & TOX21$\uparrow$ & SIDER$\uparrow$ \\ 
    \midrule
        Galactica-125M & \underline{0.8216} & \underline{0.7805} & 0.7314 & \underline{0.7453} & \underline{0.7642} \\ 
        ICMA (Galactica-125M) & \textbf{0.8941} & \textbf{0.8680} & \underline{0.7412} & \textbf{0.7476} & \textbf{0.7673} \\ 
        Mistral-7B & 0.4926 & 0.4829 & 0.5866 & 0.4386 & 0.3771 \\ 
        ICMA (Mistral-7B) & 0.7995 & 0.6775 & \textbf{0.7441} & 0.4761 & 0.4838 \\
    \bottomrule
    \end{tabular}
    }
    \label{tab:molenet}
\end{table}

\subsection{Case Study}
Here, we demonstrate the performance of different methods by introducing detailed examples in both the Cap2Mol and Mol2Cap sub-tasks. 

\begin{figure*}[htb]
    \centering
    
    \includegraphics[width=\textwidth]{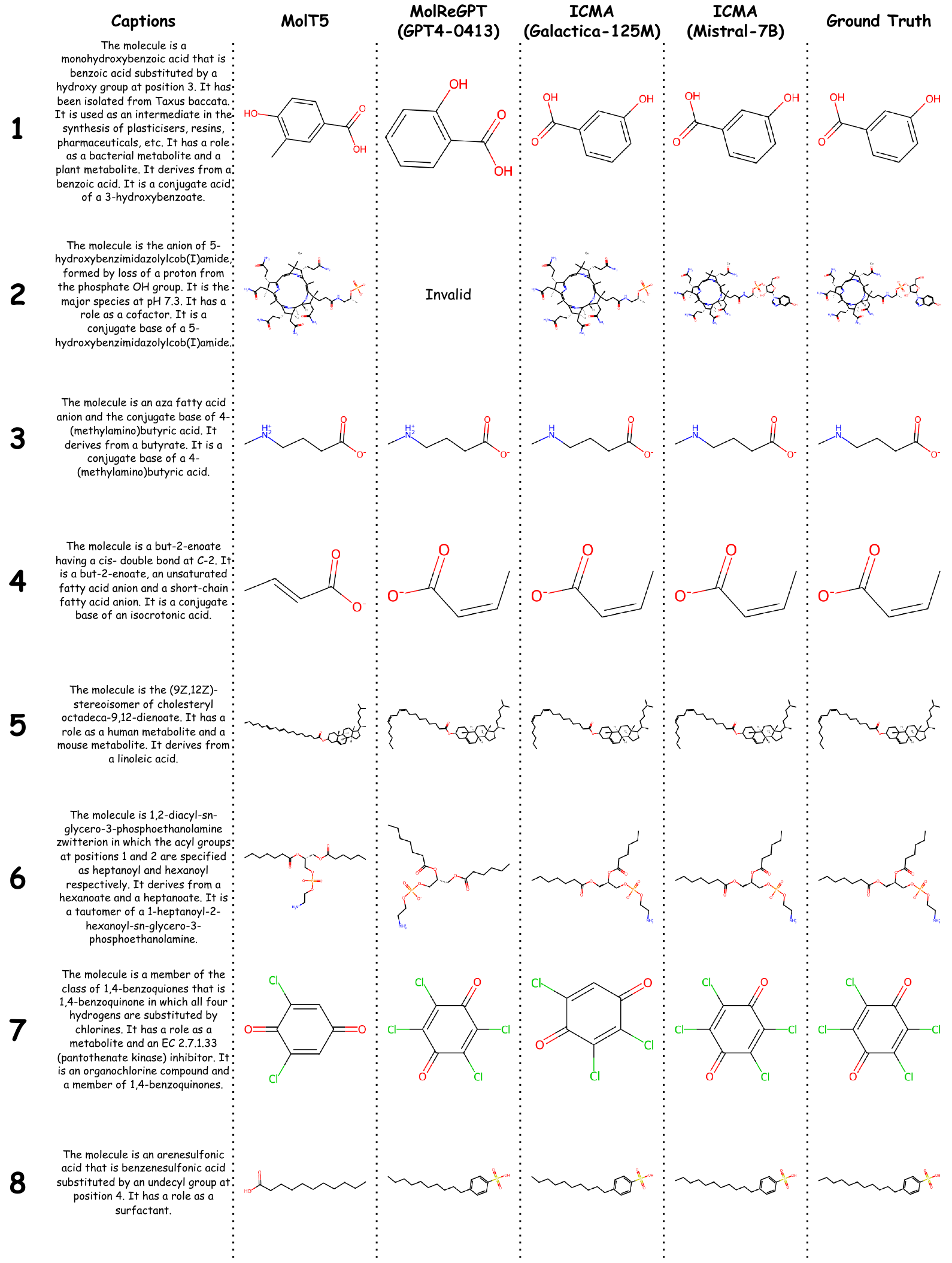}
    
    \caption{Molecule examples generated by different models based on the given captions in the Cap2Mol task.
    }
    \label{fig:case_c2m}
    
\end{figure*}

\noindent\textbf{Cap2Mol} Figure \ref{fig:case_c2m} showcases a series of molecule examples produced by various baseline methods and our proposed approach, ICMA. In the first example, both MolT5 and MolReGPT exhibited errors in the positions of functional groups, while ICMA precisely replicated the correct structure. Meanwhile, in the third example, MolT5 and MolReGPT made a mistake in the type of functional groups, generating an $[NH_2]^+$ instead of the $[NH]$. What's more, across the remaining examples, ICMA could even better match the number of carbon atoms in the SMILES representations of molecules, which consistently demonstrated superior proficiency in capturing the intricate alignment between natural language descriptions and molecule structures, thereby proving the efficacy of our proposed method.

\begin{figure*}[htb]
    \centering
    
    \includegraphics[width=\textwidth]{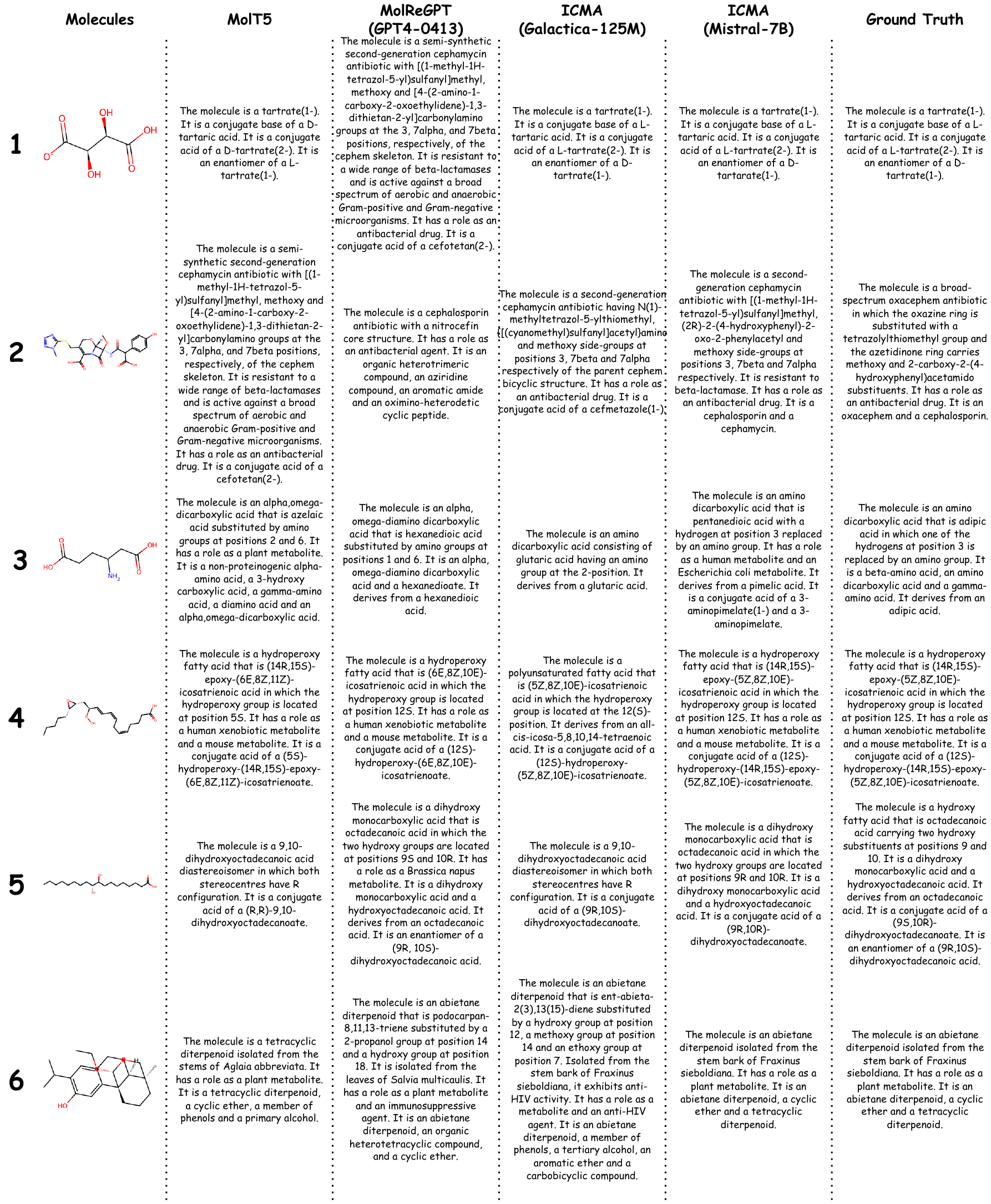}
    
    \caption{Caption examples generated by different models based on the given molecules in Mol2Cap task.
    }
    \label{fig:case_m2c}
    
\end{figure*}

\noindent\textbf{Mol2Cap} Figure \ref{fig:case_m2c} presents a set of caption examples produced by two baseline methods and our ICMA. In the first example, MolT5 incorrectly equated \emph{D-tartrate(2-)} with \emph{L-tartrate(2-)} and \emph{L-tartrate(1-)} with \emph{D-tartrate(1-)}, while MolReGPT introduced an excessive amount of irrelevant information. In contrast, ICMA could accurately empower backbone models to generate precise answers. Meanwhile, in the second example, ICMA accurately recognized that the molecule in question \emph{"has a role as an antibacterial drug"}, whereas MolReGPT failed. Moreover, in the fourth example, ICMA meticulously captured all the essential details in describing the molecule structures without a single mistake, demonstrating that contextual examples can significantly enhance the understanding of molecule structures, especially the types and positions of functional groups.

Overall, ICMA could fine-grain the detailed alignment between molecule captions and molecule SMILES representations via learning from context examples, showing better capabilities in matching the types and positions of functional groups, the number of carbon atoms, as well as the overall structures.

\section{Conclusion}
\label{sec:conclusion}
In this work, we propose In-Context Molecule Adaptation (ICMA), as a new paradigm for adapting LLMs to the molecule-caption translation task. 
Instead of domain-specific pre-training and fine-tuning, ICMA enables LLMs to utilize their in-context learning capability to learn molecule-text alignment via In-Context Molecule Tuning, which significantly improves the performance of LLMs in the molecule-caption translation task, demonstrating that LLMs are inherently in-context molecule learners.
More importantly, our study provides a viable framework for deploying advanced LLMs with billion-level parameters without extra domain alignment stages in the scientific field, making it suitable to be applied in low-data scenarios.


\section{Acknowledgement}
The research described in this paper has been partly supported by the General Research Funds from the Hong Kong Research Grants Council (project no. PolyU 15200023, and 15224524). This work is also supported by Shanghai Artificial Intelligence Laboratory (Shanghai AI Lab) and was done during Jiatong Li and Wei Liu's internship at Shanghai AI Lab.

\bibliographystyle{IEEEtran}

\bibliography{reference}
\vspace{-48pt}
{\begin{IEEEbiography}[{\includegraphics[width=1in,height=1.25in,clip,keepaspectratio]{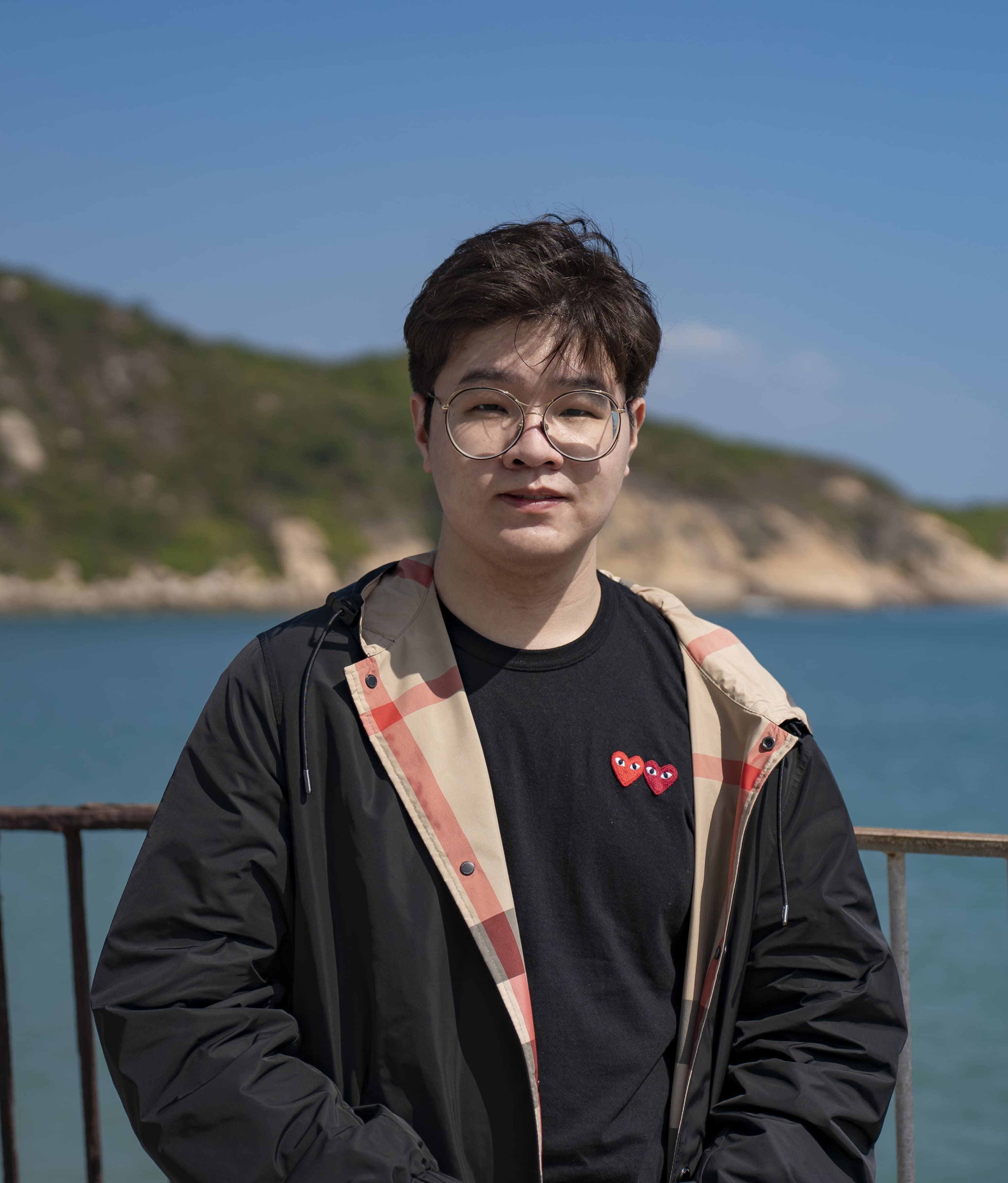}}]{Jiatong Li} is currently a PhD student of the Department of Computing (COMP), The Hong Kong Polytechnic University (funded by HKPFS). Before joining the PolyU, he received my Master's degree of Information Technology (with Distinction) from the University of Melbourne, under the supervision of Dr. Lea Frermann. In 2021, he got his bachelor's degree in Information Security from Shanghai Jiao Tong University. His interest lies in Natural Language Processing, Drug Discovery, and Recommender Systems. He has published innovative works in top-tier conferences such as IJCAI and ACL. For more information, please visit https://phenixace.github.io/.

\end{IEEEbiography}

\vspace{-24pt}
{\begin{IEEEbiography}[{\includegraphics[width=1in,height=1.25in,clip,keepaspectratio]{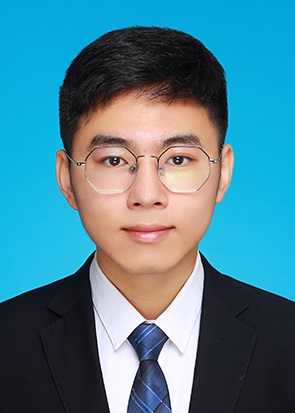}}]{Wei Liu} is currently a PhD student of the Department of Computer Science and Engineering, Shanghai Jiao Tong University, jointly trained with the Shanghai Artificial Intelligence Laboratory. In 2021, he got his bachelor's degree in Mechanical Engineering from Shanghai Jiao Tong University. His interest lies in AI-Driven Drug Design (AIDD) and Natural Language Processing. He has published works in conferences and journals such as ICLR and MBE.

\end{IEEEbiography}

\vspace{-24pt}
{\begin{IEEEbiography}[{\includegraphics[width=1in,height=1.25in,clip,keepaspectratio]{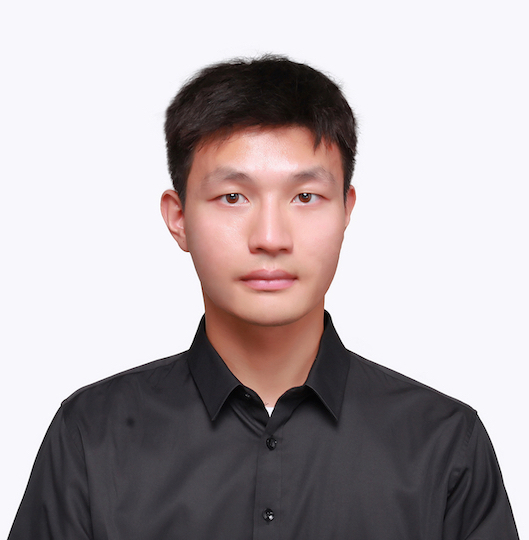}}]{Zhihao Ding} currently a PhD student of the Department of Computing (COMP), Hong Kong Polytechnic University (PolyU). Before joining the PolyU, he received his B.S. degree and M.S. degree from Northeastern University and Xi'an Jiaotong University in 2019 and 2022, respectively. His research includes Graph Neural Networks, Drug Discovery, and AI4Science. He has published innovative works in top-tier conferences such as VLDB and Neurocomputing.

\end{IEEEbiography}

\vspace{-24pt}
{\begin{IEEEbiography}[{\includegraphics[width=1in,height=1.25in,clip,keepaspectratio]{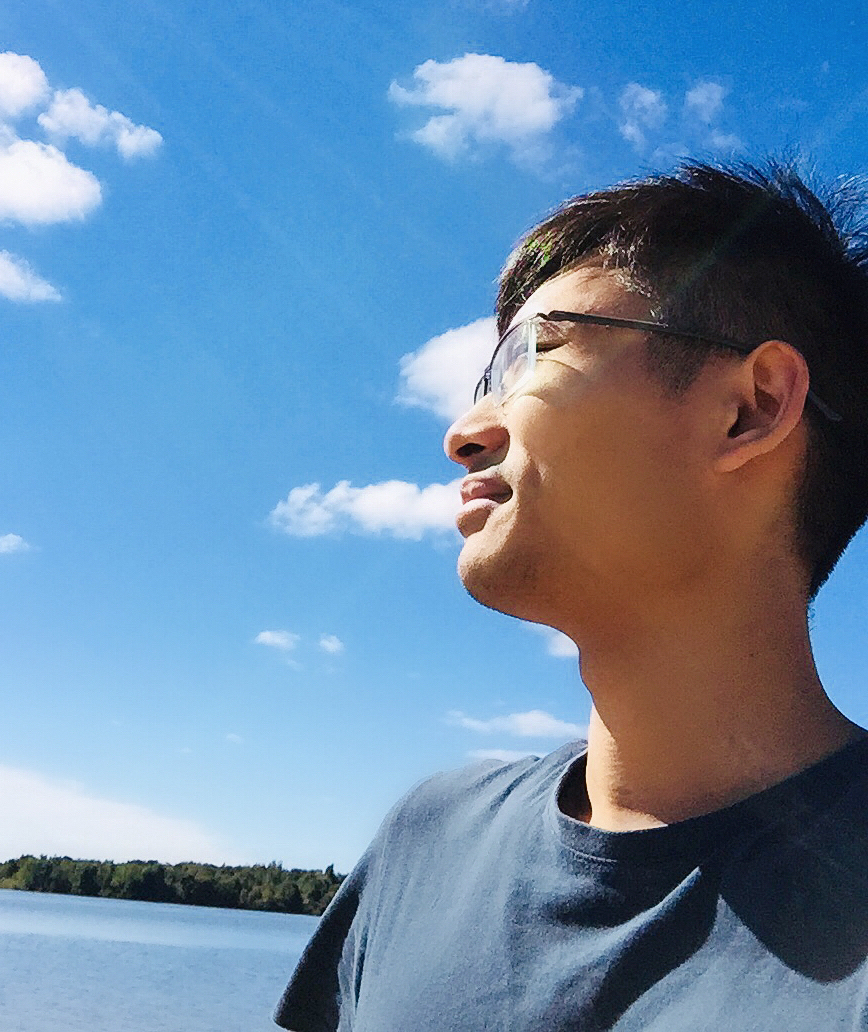}}]{Wenqi Fan} is a research assistant professor of the Department of Computing at The Hong Kong Polytechnic University (PolyU). He received his Ph.D. degree from the City University of Hong Kong (CityU) in 2020.
From 2018 to 2020, he was a visiting research scholar at Michigan State University (MSU). 
His research interests are in the broad areas of machine learning and data mining, with a particular focus on Recommender Systems, Graph Neural Networks, and Trustworthy Recommendations. He has published innovative papers in top-tier journals and conferences such as  TKDE, TIST, KDD, WWW, ICDE, NeurIPS, SIGIR, IJCAI, AAAI, RecSys, WSDM, etc. 
He serves as top-tier conference (senior) program committee members and session chairs (e.g., ICML, ICLR, NeurIPS, KDD, WWW, AAAI, IJCAI, WSDM, etc.), and journal reviewers (e.g., TKDE, TIST, TKDD, TOIS, TAI, etc.). 
More information about him can be found at https://wenqifan03.github.io.

\end{IEEEbiography}

\vspace{-24pt}
{\begin{IEEEbiography}[{\includegraphics[width=1in,height=1.25in,clip,keepaspectratio]{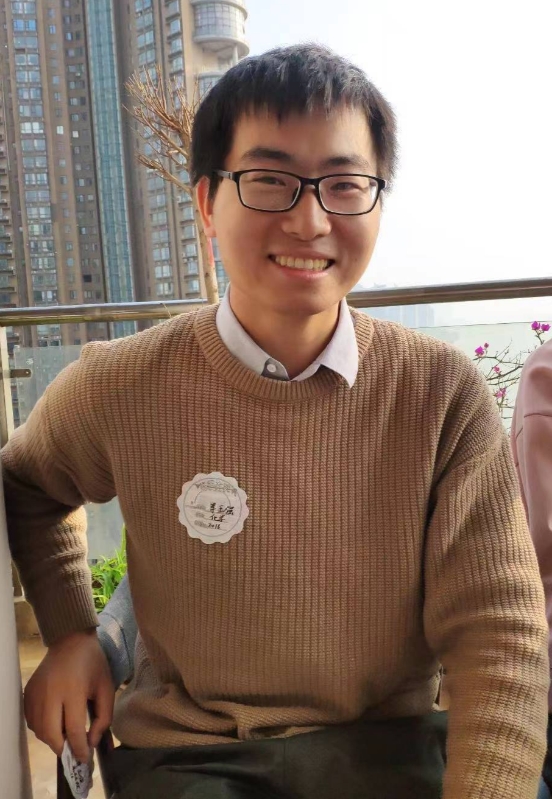}}] {Yuqiang Li} received a Bachelor of Science degree from Central South University in Changsha, China, and a Ph.D. in Chemistry from Wuhan University in Wuhan, China. He has previously held a lecturer position at Central South University. He is currently a junior researcher at the Shanghai AI Lab. His research interests encompass chemistry, machine learning, and large language models. He remains actively engaged in the scientific community, serving as a reviewer for journals such as Science Advance.

\end{IEEEbiography}

\vspace{-24pt}
\begin{IEEEbiography}
[{\includegraphics[width=1in,height=1.25in,clip,keepaspectratio]{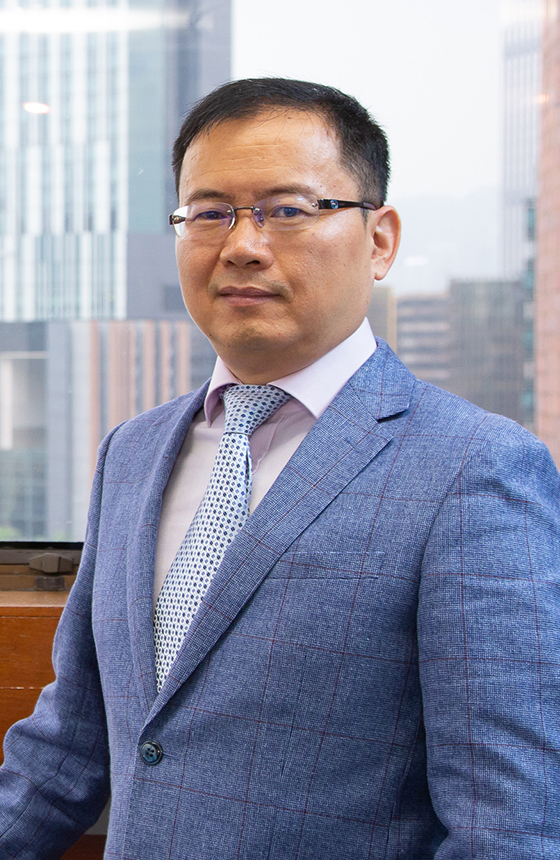}}]{Qing Li}
received the B.Eng. degree from Hunan University, Changsha, China, and the M.Sc. and Ph.D. degrees from the University of Southern California, Los Angeles, all in computer science.
He is currently a Chair Professor (Data Science) and the Head of the Department of Computing, the Hong Kong Polytechnic University. He is a Fellow of IEEE and IET, a member of ACM SIGMOD and IEEE Technical Committee on Data Engineering. 
His research interests include object modeling, multimedia databases, social media, and recommender systems. 
He has been actively involved in the research community by serving as an associate editor and reviewer for technical journals, and as an organizer/co-organizer of numerous international conferences. 
He is the chairperson of the Hong Kong Web Society, and also served/is serving as an executive committee (EXCO) member of IEEE-Hong Kong Computer Chapter and ACM Hong Kong Chapter. In addition, he serves as a councilor of the Database Society of Chinese Computer Federation (CCF), a member of the Big Data Expert Committee of CCF, and is a Steering Committee member of DASFAA, ER, ICWL, UMEDIA, and WISE Society. 
\end{IEEEbiography}
} 

\end{document}